\documentclass[journal]{IEEEtran}

\usepackage{amsmath}

\usepackage{amssymb}
\DeclareMathOperator*{\minimize}{minimize}
\DeclareMathOperator*{\argmin}{argmin}
\usepackage{cite}
\usepackage{algorithm}
\usepackage{algpseudocode}
\usepackage{amsmath}
\usepackage{setspace, etoolbox, caption}

\usepackage{tikz,xcolor,hyperref}

\definecolor{lime}{HTML}{A6CE39}
\DeclareRobustCommand{\orcidicon}{%
	\begin{tikzpicture}
	\draw[lime, fill=lime] (0,0) 
	circle [radius=0.16] 
	node[white] {{\fontfamily{qag}\selectfont \tiny ID}};
	\draw[white, fill=white] (-0.0625,0.095) 
	circle [radius=0.007];
	\end{tikzpicture}
	\hspace{-2mm}
}
\foreach \x in {A, ..., Z}{%
	\expandafter\xdef\csname orcid\x\endcsname{\noexpand\href{https://orcid.org/\csname orcidauthor\x\endcsname}{\noexpand\orcidicon}}
}

\begin{document}

\title{Accurate, Interpretable, and Fast Animation: An Iterative, Sparse, and Nonconvex Approach}

\author{Stevo~Racković\orcidA{},
        Cláudia~Soares\orcidB{},
        Dušan~Jakovetić\orcidC{}, \textit{Member IEEE},
        and~Zoranka~Desnica\orcidD{}
        \thanks{This work has received funding from the European Union's Horizon 2020 research and innovation program under the Marie Sklodowska-Curie grant agreement No 812912, and from strategic project NOVA LINCS (FCT UIDB/04516/2020). The work has also been supported in part by the Ministry of Education, Sicence and Technological Development of the Republic of Serbia (Grant No. 451-03-9/2021-14/200125).}
        \thanks{S. Racković is with the Institute for Systems and Robotics, Instituto Superior Técnico, Lisbon,
1049-001 Portugal e-mail: stevo.rackovic@tecnico.ulisboa.pt}
		\thanks{C. Soares is with the Computer Science Department, NOVA School of Science and Technology, Caparica, 2825-149 Portugal.}
		\thanks{D. Jakovetić is with the Department of Mathematics, Faculty of Sciences, University of Novi Sad, Novi Sad, 21000 Serbia.}
		\thanks{Z. Desnica is with 3Lateral Animation Studio, Epic Games Company.}
        }

\maketitle
\begin{abstract}
Digital human animation relies on high-quality 3D models of the human face—\textit{rigs}. A face rig must be accurate and, at the same time, fast to compute. One of the most common rigging models is the blendshape model. We propose a novel algorithm for solving the nonconvex inverse rig problem in facial animation. Our approach is model-based, but in contrast with previous model-based approaches, we use a quadratic instead of the linear approximation to the higher order rig model. This increases the accuracy of the solution by $8\%$ ($0.26mm$) on average and, confirmed by the empirical results, increases the sparsity of the resulting parameter vector --- an important feature for interpretability by animation artists. The proposed solution is based on a Levenberg-Marquardt (LM) algorithm, applied to a nonconvex constrained problem with sparsity regularization. In order to reduce the complexity of the iterates, a paradigm of Majorization Minimization (MM) is further invoked, which leads to an easy to solve problem that is separable in the parameters at each algorithm iteration. The algorithm is evaluated on a number of animation datasets, proprietary and open-source, and the results indicate the superiority of our method compared to the standard approach based on the linear rig approximation. Although our algorithm targets the specific problem, it might have additional signal processing applications.
\end{abstract}


\IEEEpeerreviewmaketitle

\section{Introduction}

\IEEEPARstart{M}{odern} facial animation demands ever more realistic and complex characters to satisfy user needs and avoid the uncanny valley effect \cite{mori2012uncanny}. The underlying functions of the animation models include high-level non-linear terms, increasing the complexity of the animation task. Manually animating such a model is expensive in terms of both time and effort, hence the automated solutions are constantly being developed \cite{choe2001analysis, li2010example, joshi2006learning, yu2014regression, ccetinaslan2016position}. One promising direction is to apply machine learning, e.g., neural networks \cite{holden2015learning, holden2016learning}, but to train a good model, one needs a large amount of training data, which is still expensive to obtain. For this reason, we focus on developing a model-based optimization method for solving the inverse rig problem \cite{holden2015learning} in blendshape animation. It is desirable that the algorithm is parallelizable in order to reduce the fitting time and eventually lead to a real-time animation solution. Also, it should take into account a number of domain-specific constraints. The optimization literature offers a number of efficient algorithms applicable to a range of non-linear problems \cite{boyd_2011, cao_2017, Notarnicola2018, jakovetic2020primal}, but instead of just applying an off-the-shelf model, we adjust the algorithms and build a customized solution that will target the above mentioned domain-specific constraint.

\begin{figure}
    \centering
    \includegraphics[width=\linewidth]{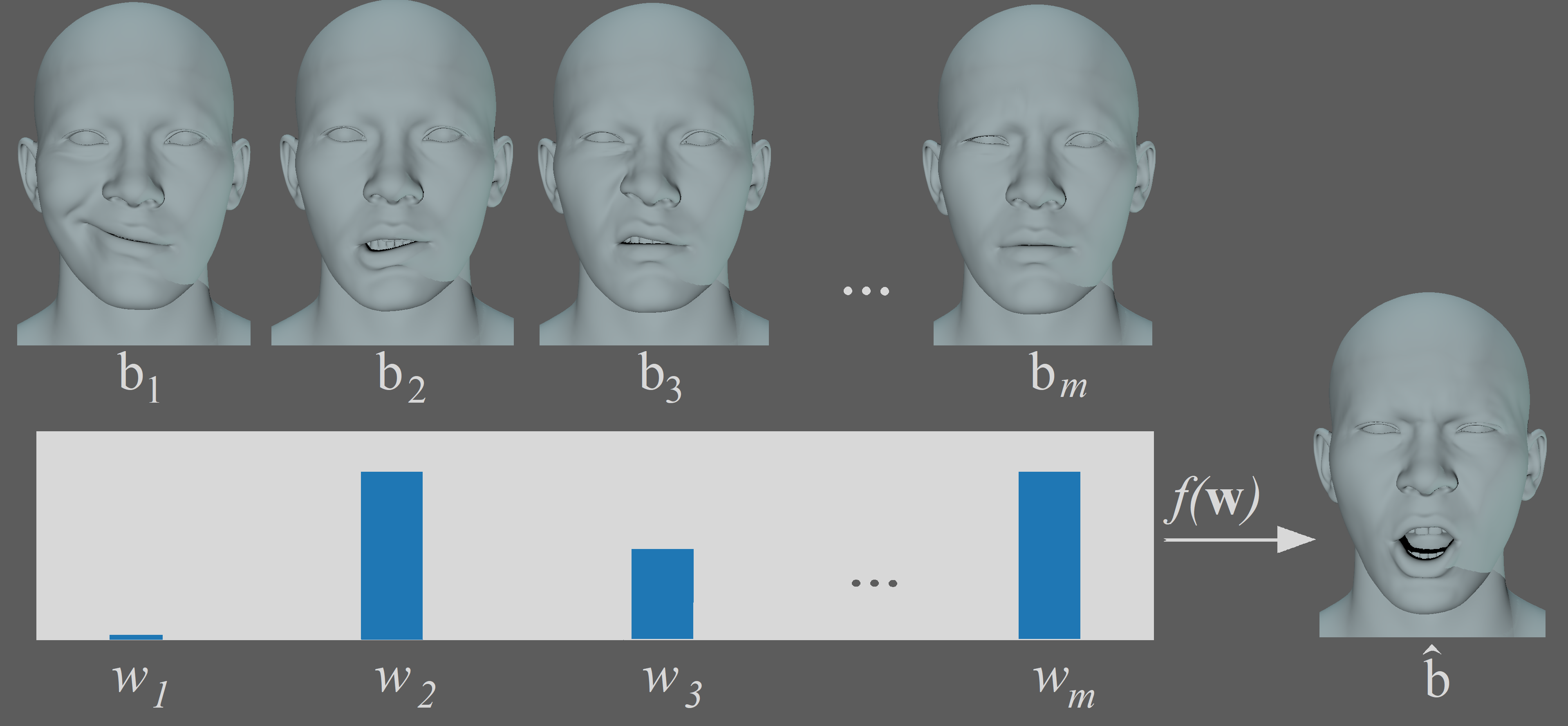}
    \caption{Face meshes in the top of the figure represent blendshapes $\textbf{b}_1,...,\textbf{b}_m$ of a character, the bars bellow are activation weights $\textit{w}_1,...,\textit{w}_m$ corresponding to each of the blendshapes and the estimated face mesh $\hat{\textbf{b}}$ is obtained by applying a rig function $f(\textbf{w})$.}
    \label{fig:blendshape_model}
\end{figure}

The functions used in the facial animation are blendshape rigs \cite{lewis2014practice}. Traditionally, blendshape functions are linear $f(\textbf{w}) = \textbf{Bw}$ \cite{alkawaz2015blend}, and the inverse rig problem migh be posed as 
\begin{equation}\label{eq:obj_0}
    \minimize_{\textbf{w}}\|\textbf{Bw}-\hat{\textbf{b}}\|^2
\end{equation}
where $\textbf{w}\in\mathbb{R}^m$ is a vector of parameters, or controller weights, $\textbf{B}\in\mathbb{R}^{3n\times m}$ is a blendshape matrix, whose columns are blendshape vectors $\textbf{b}_1,...,\textbf{b}_m\in\mathbb{R}^{3n}$, and $\hat{\textbf{b}}\in\mathbb{R}^{3n}$ is a given face mesh as depicted in Figure \ref{fig:blendshape_model}. For this simple case (with possible constraints or regularization terms) solutions already exist \cite{ccetinaslan2016position,boyd2004convex}, but modern blendshape models are increasingly complex and non-linear. Hence, more complex solutions are needed in order to keep a high-level accuracy of the solution. State-of-the-art blendshape models add several layers of corrections in order to increase the expressivity of a character and make local deformations more realistic. This yields rig functions that are typically third or fourth-order polynomials (see Section \ref{section_iii}).

\subsection{Contributions} 

In this paper, we propose a novel algorithm for solving the inverse rig problem for facial animation. We approach this problem by approximating a complex blendshape rig with a second-order polynomial form\footnote{Blendshape models are traditionally linear, but in modern animation several levels of correction might be added on top of a linear model, leading to more accurate local representation of the face. Second-order approximation means that we consider only the first level of the correction (see Section \ref{proposed_modeling}).} and construct an iterative procedure with constraints on the range of the parameters. Our model is employing the Levenberg-Marquardt algorithm \cite{Levenberg1944AMF, marquardt1963algorithm} in the light of the Majorization Minimization framework \cite{becker1997algorithms, lange2000optimization, hunter2004tutorial, zhang2007surrogate}, and produces an accurate and sparse solution. The proposed method is designed such that the major computational step is parallelizable component-wise with respect to the components of the vector of controller weights, and hence is amenable to complex models and real-time animation.

At each iteration we solve a scalar quartic equation per component, that offers a fast to compute solution. Instead of using all the corrective levels of the rig, as in (\ref{eq:rig_function}), we approximate the rig with a second-order model since this is a complexity level that allows us to model the inverse rig solution efficiently, and yet it is considerably more accurate than the linear approximation (see Section \ref{section_iii}). We estimate the inverse rig solution through the following constrained problem:
\begin{equation}\label{eq:objective_1}
    \minimize_{\textbf{0}\leq \textbf{w}\leq \textbf{1}}\|f(\textbf{w})-\hat{\textbf{b}}\|^2 + \lambda \textbf{1}^T\textbf{w}
\end{equation}
where $\lambda\geq 0$ is a regularization parameter. Besides the choice of a rig function, the objective function (\ref{eq:objective_1}) differs from (\ref{eq:obj_0}) in the regularization term that forces a solution to be sparse --- which is important in animation because sparser vectors are more likely to produce semantically correct activations, and are easier to manually adjust (if needed).

The proposed algorithm is based on the application of the Levenberg-Marquardt (LM) algorithm \cite{kelley1999iterative,boyd2004convex} for minimizing a constrained problem (\ref{eq:objective_1}). LM is viewed as a trust region algorithm \cite{yuan2000review} which aligns well with the structure of our problem (tight box constraints), and together with its good convergence properties \cite{ranganathan2004levenberg,pujol2007solution} it imposes as a natural choice for solving the inverse rig. However, an LM iterate is still complex (even under the quadratic approximation of the rig), hence we devise an upper bound to the objective that is easier to minimize, and apply the Majorization Minimization (MM) \cite{becker1997algorithms, lange2000optimization, hunter2004tutorial, zhang2007surrogate} to solve the problem. The upper bound consists of a set of simple and component-wise independent equations. Experiments indicate that our algorithm leads to a more accurate solution than the standard linear approximation while keeping the estimated vector sparse. Linear model has shorter computational time, but we build on top of linear solution and achieve the convergence in only a few additional iterations.

We next provide a more detailed literature review to help us contrast our contributions with respect to existing work. We organize the literature review in three different categories. Subsection \ref{lierature_animation} overviews the field of facial animation, and more concretely, contrasts our work with respect to other inverse rig solutions. Subsections \ref{lierature_lm} and \ref{literature_mm} respectively review LM and MM methods that are the key technical ingredients that we need in order to develop the proposed inverse rig method.

\subsection{Literature Review} 

\subsubsection{Facial Animation}\label{lierature_animation}
There is no original reference paper for the blendshape model, but a thorough introduction can be found in \cite{lewis2014practice, choe2001analysis, li2010example, lewis2010direct}. The main components of a model are a neutral face mesh, a blendshape basis (local deformations of a neutral face), and a set of controllers corresponding to the meshes from the basis. Creation of the blendshape basis is discussed in \cite{li2010example, li2013realtime, neumann2013sparse, lewis2014practice}.

The next stage in the pipeline is adjusting activation weights to produce a desired animation---this is called the inverse rig problem and represents the main bottleneck in production due to the time involved. This is also the main focus of our paper, where we want to develop a solution that reduces execution time and complexity in solving inverse rig while working with accurate models. Automated solutions for certain linear forms of the rig are proposed in \cite{choe2001analysis, li2010example, joshi2006learning, yu2014regression, ccetinaslan2016position}. In order to enhance the fidelity of expression, additional corrective blendshapes are introduced, as explained in (\ref{eq:rig_function}), yielding a nonconvex problem---as studied in \cite{holden2016learning, song2017sparse}. One generalization of inverse rig learning is the problem of direct manipulation that considers an interface allowing a user to drag vertices of the face directly in order to produce the desired expression \cite{lewis2010direct}. 

A possible approach towards a distributed inverse rig solvers is via using a face segmentation or clustering. It allows different regions of the face to get observed and processed independently or in parallel. Early works consider a simple split of the face into upper and lower sets of markers \cite{choe2001analysis}. Later, these models are sought to be automatic \cite{joshi2006learning,song2017sparse} or semi-automatic \cite{na2011local, tena2011interactive, fratarcangeli39fast}. All these works use either topological positions of vertices in the face or their correlation over animation sequence, neglecting the underlying blendshape model. Clustering based on the underlying deformation model has been considered in \cite{romeo2020data} and \cite{rackovic}, where a goal of the former was to add a secondary motion to an animated character, and the latter proposes a segmentation for solving the inverse rig locally in a distributed fashion. However, in this paper we do not investigate further application of face clusters in solving our objective.
 
 In summary, the approaches to solving the inverse rig problem can be divided into data-based and model-based. Data-based solutions are popular due to their ability to provide accurate solutions even for complex rig functions \cite{holden2015learning, holden2016learning, neumann2013sparse}. However, the data acquisition is too expensive, which is why in this paper we consider a model-based approach. In this case, the literature examines only the linear rigs \cite{choe2001analysis,sifakis2005automatic,ccetinaslan2016position,li2010example}, yielding convex optimization problems. We propose using a quadratic rig, that yields a non-convex objective, but (as confirmed in our experiments) leads to a more accurate and sparser solution.

 \subsubsection{Levenberg-Marquardt Algorithm}\label{lierature_lm}
The Levenberg-Marquardt (LM) \cite{kelley1999iterative, boyd2004convex, ranganathan2004levenberg} algorithm was originally proposed as a solution to non-linear least squares problems \cite{Levenberg1944AMF, marquardt1963algorithm}. It blends two well known algorithms, Gradient Descent (GD) \cite{kelley1999iterative,boyd2004convex,ruder2016overview} and Gauss-Newton (GN) \cite{kelley1999iterative,boyd2004convex,wang2012gauss}, in order to exploit the benefits of both --- strong convergence properties of GD and a local quadratic convergence rate of GN \cite{li2019convergence, 7131516,dennis1996numerical,siregar2018analysis,yamashita2001rate,fan2005quadratic,ahookhosh2019local}. The objective function of LM may be equivalently presented as a constrained optimization problem, which further gave rise to Trust Region algorithms \cite{yuan1994trust,conn2000trust,berghen2004levenberg}. A modern LM is viewed as Gauss-Newton using a trust region approach \cite{yuan2000review}. 

Due to its favorable convergence properties, LM became a popular algorithm for solving optimization problems. Intuitive explanations of the algorithm and its variants are given in \cite{ranganathan2004levenberg,pujol2007solution}. There is a long line of research that provides specific applications and implementations of LM, especially in the framework of Neural Networks \cite{ngia2000efficient, asirvadam2002parallel, auger2012making, nguyen2013combining, yu2018levenberg, gavin2019levenberg, sarkka2020levenberg}.

Nonconvex problems arise often in signal processing \cite{cheung2006constrained, lopes_2008, keaktos_2011, zhu_2011, sahu_2016}, and recently a number of papers have proposed problem-specific adaptations of LM for solving them \cite{fu_2015,xu_2016,meng_2019,ok_2018}. To the best of our knowledge, LM has not been applied in the facial animation literature. In this paper we develop a model for solving the inverse rig in animation, based on LM, that efficiently solves a constrained non-linear least squares problem with sparsity regularization and box constraints. 

 \subsubsection{Majorization Minimization}\label{literature_mm}
 Majorization Minimization (MM) \cite{becker1997algorithms, lange2000optimization, hunter2004tutorial, zhang2007surrogate} is a generalization of the expectation maximization algorithm \cite{dempster1977maximum, jacobson2007expanded}. The idea of the algorithm is to iteratively substitute the original objective function with surrogate functions that are easier to minimize and bound the original objective from above, while touching the objective in the previous iterate. Depending on the problem at hand, different assumptions on the majorizer are often made, e.g., the function being continuously differentiable, convex, Lipschitz smooth, etc. \cite{lange1995gradient, schifano2010majorization, chouzenoux2017stochastic, marnissi2020majorize, lange2021nonconvex, fest2021stochastic} in order to guarantee the convergence. 
Reference \cite{hunter2004tutorial} offers a good tutorial that covers different types of problems solved via MM, with extensive literature review, while \cite{sun2016majorization} provides a number of example applications in signal processing, communications, and machine learning.
 
As mentioned in \cite{sun2016majorization}, it is desirable for a surrogate function to be separable in variables, convex and smooth. In this paper, we use a majorization function with a closed-form solution that is separable in components. The powerful MM technique and a judicious formulation of tight upper bounds allowed us to derive an efficient LM-type method for solving the specific non-linear least squares constrained problem relevant to inverse rig as explained above. 

\subsection{Notation} 

We denote by $\mathbb{R}$ the set of real numbers and by $\mathbb{R}^m$ the real Euclidean space of dimension $m$.
Scalars, vectors and matrices are denoted by lowercase ($a$), bold lowercase ($\textbf{a}$) and bold uppercase ($\textbf{A}$) letter respectively. We use subscripts to denote elements of vectors and rows/columns of matrices: $a_i$ is the $i^{th}$ element of a vector $\textbf{a}$, $\textbf{A}_i$ is $i^{th}$ row of a matrix $\textbf{A}$, $\textbf{A}_{:i}$ is $i^{th}$ column of a matrix $\textbf{A}$ and $A_{ij}$ is the element of a matrix $\textbf{A}$ in the row $i$ and column $j$. The symbol $\textbf{0}$ ($\textbf{1}$) represents vector with all elements equal to $0$ ($1$). In case of vectors, the inequality operators ($<,\leq,\geq,>$) are considered to be element-wise. $\|\cdot\|$ represents the $l_2$ norm and $\|\cdot\|_{\infty}$ represents the $l_{\infty}$ norm. Functions are represented using standard notation, either by Latin or Greek letters, but always with their argument indicated, e.g.: $f(x),\,\phi(x),\,F(x),\,\Phi(x).$ When we discuss iterative procedures we use notation $(t)$ to represent a specific iteration as subscript, e.g., $\mathbf{a}_{(t)}$ is the iterate at $t$-th iteration. Operator $\dagger$ represents the pseodoinverse, i.e. $\textbf{A}^{\dagger} = (\textbf{A}^T\textbf{A})^{-1}\textbf{A}^T$. The mapping $P_{[a,b]}(x):\mathbb{R}\rightarrow [a,b]$ is the projection of a scalar variable $x$ onto the interval $[a,b]$. 
The largest singular value of a matrix $\textbf{A}$ is denoted as $\sigma_{\text{max}}(\textbf{A})$, and the largest and the smallest eigenvalues of $\textbf{A}$ are $\lambda_{\text{max}}(\textbf{A})$ and $\lambda_{\text{min}}(\textbf{A})$ respectively.

This paper is organized as follows. Section \ref{section_preliminaries} covers the theoretical concepts needed for formulating the inverse rig problem and describing the proposed algorithm. Section \ref{section_iii} formulates the problem and gives a derivation of the proposed algorithm. Section \ref{section_results} shows the results of the numerical experiments performed on several animation datasets, and finally Section \ref{section_conclusion} concludes the paper with a discussion.


\section{Preliminaries}\label{section_preliminaries}

 Section \ref{preliminaries_rig} explains rig-based modeling in facial animation and defines the inverse rig problem. Section \ref{preliminaries_lm} covers background on the Levenberg-Marquardt (LM) algorithm and how it is applied for solving the inverse rig, and Section \ref{preliminaries_lm} describes the Majorization Minimization approach and how it is used further to simplify the iterates of LM in the proposed algorithm.

\subsection{Rig Function and Inverse Rig}\label{preliminaries_rig} 

A rig function in animation is a mapping $f(\textbf{w}):\mathbb{R}^{m}\rightarrow\mathbb{R}^{3n}$ that takes a set of $m$ controller parameter values $\textbf{w}$ and deforms a character mesh $\hat{\textbf{b}}\in\mathbb{R}^{3n}$ in space, according to the space of motion defined via controllers. A particular class of rig functions that are of interest to us are blendshape-based rigs \cite{lewis2014practice}. In its simplest form, the blendshape rig represents an affine transformation of the parameters and, hence, it is easy to understand and apply. In this section we present the main principles of blendshape animation for a human face.

\begin{figure}
    \centering
    \includegraphics[width=\linewidth]{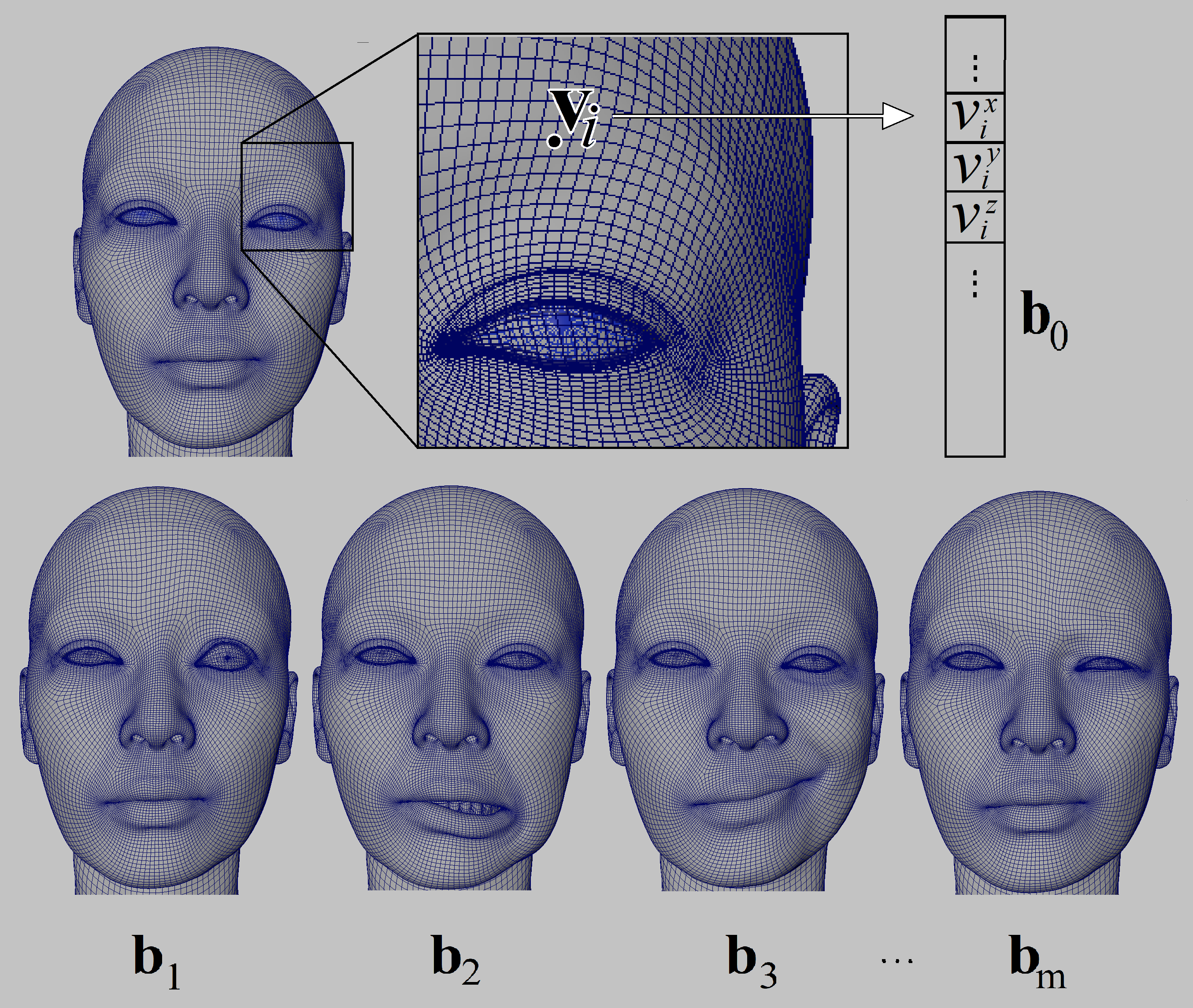}
    \caption{Vectorization of a face mesh $\textbf{b}_0$ (neutral face mesh on the top) and blendshapes $\textbf{b}_1$,...,$\textbf{b}_m$. Each vertex $\textbf{v}_i$ from the mesh is separated by components ($x,y,z$) and concatanated into a vector $\textbf{b}_0$.  (See section \ref{preliminaries_rig})}
    \label{fig:vectorization}
\end{figure}

Consider a 3D face model in the neutral position, i.e., no muscle activations or any visible expression, as depicted in Figure \ref{fig:vectorization}. It consists of $n$ vertices $\textbf{v}_1,...,\textbf{v}_n\in\mathbb{R}^{3}$ on the surface mesh. We unravel $x,y,z$ coordinates of each vertex $\textbf{v}_i$ and stack them into a single vector $\textbf{b}_0\in\mathbb{R}^{3n}$ such that $\textbf{b}_0 = [v_1^x,v_1^y,v_1^z,...,v_n^x,v_n^y, v_n^z]^T$ (see Figure \ref{fig:vectorization}). Additionally, we have a set of $m$ topologically identical copies of the neutral face (none of the vertices or edges can be added or removed, we are only allowed to move them in space), but each with a local deformation. These meshes are called blendshapes, and each of them represents an atomic facial deformation, and combining them produces more complex expressions (and it spans the space of possible deformations). Each of these meshes is vectorized in the same manner as a neutral mesh, and we obtain $m$ vectors $\textbf{b}_1,...,\textbf{b}_m\in\mathbb{R}^{3n}$. A blendshape matrix $\textbf{B}\in\mathbb{R}^{3n\times m }$ is then formed as a matrix whose columns are blendshape vectors $\textbf{B}:=[\textbf{b}_1,...,\textbf{b}_m]$. This procedure is illustrated in Figure \ref{fig:vectorization}.

In the linear blendshape model, any feasible facial expression can be obtained as
$$f(\textbf{w}) = \textbf{b}_0 + \textbf{B}\textbf{w}$$ where $\textbf{w}=[w_1,...,w_m]^T$ is a vector of activation weights for each blendshape. The mapping $f(\textbf{w}):\mathbb{R}^m \rightarrow \mathbb{R}^{3n}$, from parameters $\textbf{w}$ into a mesh space, is called a rig.

In modern animation, we work with hundreds of blendshapes in the basis and thousands of face vertices. Due to this high dimensionality, artifacts in the face mesh are common. It happens that a pair of blendshapes $\textbf{b}_i$ and $\textbf{b}_j$, when activated simultaneously, produces a local deformation that is different from what the artist expected. In that case, the artist sculpts a new mesh $\hat{\textbf{b}}$, that is the desired output for this pair, and extracts the difference between the obtained and desired meshes as a corrective blendshape $\textbf{b}^{\{i,j\}}$:
$$\textbf{b}^{\{i,j\}} = \hat{\textbf{b}} - (\textbf{b}_0+\textbf{b}_i+\textbf{b}_j).$$

Now, whenever the blendshapes $\textbf{b}_i$ and $\textbf{b}_j$ are activated simultaneously, the corrective blendshape $\textbf{b}^{\{i,j\}}$ is activated as well, by a coefficient that is equal to the product of the two coefficients:
$$f(w_i,w_j) = \textbf{b}_0 + w_i\textbf{b}_i + w_j\textbf{b}_j + w_iw_j\textbf{b}^{\{i,j\}}.$$

Using the same reasoning, we can have the higher order corrective terms as well ---  $\textbf{b}^{\{i,j,k\}}$ for triplets of controllers (corresponding to products of weights $w_iw_jw_k$), $\textbf{b}^{\{i,j,k,l\}}$ for set of four, and so on\footnote{Multiplying the corrective blendshapes by the products of the individual weights $w_i$'s that correspond to the corrective blendshape has been empirically shown itself to lead to semantically accurate models while at the same time not extending the dimensionality of the weight vector $\textbf{w}$ to be designed.}. 

Even tough the creation of the blendshape base is a work intensive task, once we have all the blendshapes, the rig function is simple and straightforward to use. There is a more complex and common problem that is called an inverse rig problem, and solving it is our main objective in this paper. The inverse rig considers a reference mesh $\hat{\textbf{b}}\in\mathbb{R}^{3n}$ that is conventionally obtained as a 3D scan of an actor, and the task is to find an optimal estimate of the controller vector $\hat{\textbf{w}}$ so that $f(\hat{\textbf{w}})\approx \hat{\textbf{b}}$. The problem is often stated as a least squares minimization: 
$$\minimize_{\textbf{w}} \|f(\textbf{w})-\hat{\textbf{b}}\|_2^2$$
with possible constraints on the structure or sparsity of $\textbf{w}$.

\subsection{Levenberg-Marquardt Algorithm}\label{preliminaries_lm} 

The Levenberg-Marquardt (LM) algorithm is a well known iterative algorithm that interpolates between gradient descent (GD) and Gauss-Newton (GN) method, overcoming the downsides of both. The primary problem that LM tries to solve is the least squares curve fitting. If there is a set of $n$ empirical pairs $(x_i,y_i)$ and a model curve is $f(x_i;\textbf{w})$, where $\textbf{w}\in\mathbb{R}^m$ are model parameters to be estimated, the residuals are defined as  $s(x_i,y_i,\textbf{w}) := f(x_i,\textbf{w})-y_i$. The corresponding optimization problem is

\begin{equation}\label{eq:cost}
    \minimize_{\textbf{w}}\, S(\textbf{w}):=\sum_{i=1}^n s(x_i,y_i,\textbf{w})^2.
\end{equation}

When minimizing (\ref{eq:cost}) via gradient descent, the iteration step, for iteration $t$, is
\begin{equation}
    \textbf{w}_{(t+1)} = \textbf{w}_{(t)} - \alpha\triangledown S(\textbf{w}_{(t)})
\end{equation}
where $\alpha>0$ is a step size and $\triangledown S(\textbf{w}):\mathbb{R}^m\rightarrow\mathbb{R}^m$ is a gradient of the cost function. This update makes small steps in the regions of a low gradient, so the algorithm's convergence is very slow. Another problem is that GD follows an oscillatory path along narrow valleys, additionally increasing the number of iterations needed for the convergence. GN uses curvature information, so it improves in this respect and exhibits a quadratic convergence rate in the vicinity of the solution. The iteration step for GN is 
\begin{equation}
    \textbf{w}_{(t+1)} = \textbf{w}_{(t)} - (\triangledown^2S(\textbf{w}_{(t)}))^{-1}\triangledown S(\textbf{w}_{(t)})
\end{equation}
where $\triangledown^2 S(\textbf{w}):\mathbb{R}^m\rightarrow\mathbb{R}^{m\times m}$ is a Hessian of the cost function (or an approximation to the Hessian).

Unfortunately, this method is susceptible to initialization, and in general, does not give a convergence guarantee. The problem with GN is that it can produce arbitrary large increments, which makes the algorithm diverge in some cases, even in the convex setting. To overcome this, the LM algorithm includes a damping parameter $\lambda>0$ that prevents too large steps:

\begin{equation}\label{eq:LM_iteration}
    \textbf{w}_{(t+1)} = \textbf{w}_{(t)} - (\triangledown^2S(\textbf{w}_{(t)}) + \lambda \textbf{I})^{-1}\triangledown S(\textbf{w}_{(t)}).
\end{equation}
LM iteration (\ref{eq:LM_iteration}) is a blend of GD and GN iterations, in the sense that small values of $\lambda$ produce a GN-like step, while very large values of $\lambda$ will make the first term inside of the brackets (\ref{eq:LM_iteration}) negligible so that the update resembles that of GD. Parameter $\lambda$ is updated over iterations --- if the objective is decreased sufficiently, $\lambda$ is reduced, but if the objective increases, $\lambda$ gets increased as well. This leads to an algorithm that is robust w.r.t. initial point, and yet it has a quadratic convergence rate in the vicinity of a solution \cite{yamashita2001rate,fan2005quadratic,ahookhosh2019local}.

LM can equivalently be presented in a framework of a trust-region optimization \cite{conn2000trust,berghen2004levenberg}, and this is the formulation that we also use in derivation of the proposed algorithm for solving the inverse rig (see ahead Algorithm \ref{alg:cap}). In this perspective, objective function in (\ref{eq:cost}) is (usually) approximated by a Taylor expansion around $\textbf{w}_{(t)}$:
$$S(\textbf{w}_{(t)}+\textbf{v}) \approx S(\textbf{w}_{(t)})+\triangledown S(\textbf{w}_{(t)})^T\textbf{v} + \frac{1}{2}\textbf{v}^T\triangledown^2S(\textbf{w}_{(t)})\textbf{v},$$
and we look for an increment vector $\textbf{v}\in\textbf{R}^m$, constrained by the radius $\Delta$ ($|v_i|<\Delta$ for $i=1,...,m$), that minimizes this approximation. The update rule (\ref{eq:LM_iteration}) is restated as 
\begin{equation}\label{eq:lm_iter}
\begin{split}
    \hat{\textbf{v}} \in \argmin_{\|\textbf{v}\|_{\infty}<\Delta}\,\,  & S(\textbf{w}_{(t)})+\triangledown S(\textbf{w}_{(t)})^T\textbf{v} + \frac{1}{2}\textbf{v}^T\triangledown^2S(\textbf{w}_{(t)})\textbf{v}\\
    \textbf{w}_{(t+1)}  & = \textbf{w}_{(t)} + \hat{\textbf{v}}.
\end{split}
\end{equation}
This is the formulation of LM that we use to construct the algorithm proposed in this paper. The main difference in this respect is the bound over an increment vector $\textbf{v}$. In all our models we have a constraint on the parameters $\textbf{0}\leq\textbf{w}\leq\textbf{1}$, hence we do not use a ball of radius $\Delta$ to limit the increment $\textbf{v}$, but rather we demand that $\textbf{0}\leq\textbf{w}_{(t)} + \textbf{v}\leq \textbf{1}$ at each iteration $t$ (see ahead Section \ref{first_quadratic}, (\ref{eq:derivation_lm_tr})).

\subsection{Majorization Minimization}\label{preliminaries_mm} 

The Majorization Minimization (MM) algorithm is a robust iterative procedure for numerical optimization. Instead of minimizing the original objective function, which might be too complex, it takes a more straightforward surrogate function and minimizes it instead. Consider that the original optimization problem is 
\begin{equation}
    \minimize_{\textbf{w}}S(\textbf{w})
\end{equation}
with a function $S(\textbf{w})$ that cannot be minimized efficiently. The idea of MM is that, at each iteration $t$, we construct an upper bound (or a surrogate) function $\Theta(\textbf{w}|\textbf{w}_{(t)})$. Here, $\textbf{w}$ is the argument of the function, and $\textbf{w}_{(t)}$ in $\Theta(\cdot|\textbf{w}_{(t)})$ designates that one surrogate function $\Theta(\cdot|\textbf{w}_{(t)})$ is associated to each iterate $\textbf{w}_{(t)}$, for any $t$, where the surrogate functions for different $t$'s may be mutually different. This function is a majorizer of the original function at point $\textbf{w}_{(t)}$, which means that it is above the original function at any point, and that the two are equal at $\textbf{w}=\textbf{w}_{(t)}$:
\begin{equation}\label{eq:majorizer}
\begin{split}
    \Theta(\textbf{w}|\textbf{w}_{(t)}) & \geq S(\textbf{w}) \text{  for any  }\textbf{w}\\
    \Theta(\textbf{w}_{(t)}|\textbf{w}_{(t)}) & = S(\textbf{w}_{(t)}).
\end{split}
\end{equation}

The majorizer is often (but not exclusively) constructed based on the Jensen's inequality, Cauchy–Schwarz inequality, or Taylor's expansion of the objective $S(\textbf{w})$. Depending on the application, a different strategy might be considered --- in our algorithm we use the Cauchy–Schwarz inequality multiple times to derive a surrogate. Irrespective of the derivation, a surrogate is sought to be simple and easy to minimize, preferably offering a closed-form solution. Once this upper bound is defined, an update rule of the MM algorithm at iteration $t$ is
\begin{equation}
    \begin{split}
        \textbf{w}_{(t+1)} \in \argmin_{\textbf{w}} \Theta(\textbf{w}|\textbf{w}_{(t)}).
    \end{split}
\end{equation}

From (\ref{eq:majorizer}) it is easy to see that this procedure decreases the objective monotonically:
$$ S(\textbf{w}_{(t)}) = \Theta(\textbf{w}_{(t)}|\textbf{w}_{(t)}) \geq \Theta(\textbf{w}_{(t+1)}|\textbf{w}_{(t)}) \geq S(\textbf{w}_{(t+1)}).$$

In the algorithm proposed here (see ahead Algorithm \ref{alg:cap}), we derive a surrogate function (upon an LM iterate) that is separable in components and offers a closed-form solution. The bound produced with this choice is tight, and the experiments show a steady decrease in the objective. 


\section{Problem Formulation}\label{section_iii}

Recall that the inverse rig problem assumes that there is a target mesh $\hat{\textbf{b}}\in\mathbb{R}^{3n}$ (e.g. a 3D scan of a face) and the task is to estimate an optimal vector of controller activation weights $\hat{\textbf{w}}\in\mathbb{R}^m$ so that the target mesh is well approximated by the rig function $f(\hat{\textbf{w}})$. The entries of the vector are constrained to $0\leq \hat{w}_i \leq 1$ for $i=1,...,m$ by the construction of a blendshape model (and the software \texttt{Autodesk Maya} will not accept values outside of this range). It is further desirable that the estimated vector $\hat{\textbf{w}}$ is sparse. There are two principal reasons for this. First, since the solution is not unique, a sparser solution is more likely to represent semantically correct activations vector. This is further going to influence the smoothness of the animation when the model transitions over the frames. Second, animators often need to understand and adjust already estimated controllers, which is increasingly harder when a large number of controllers is activated. 

In order to account for the modeling requirements above, we formulate the inverse rig problem as
\begin{equation}\label{eq:objective}
\minimize_{0\leq\textbf{w}\leq1}\| f(\textbf{w}) - \hat{\textbf{b}}\|^2 + \lambda\textbf{1}^T\textbf{w}
\end{equation}
where $\lambda\geq0$ is a regularization parameter. In our datasets the rig function has 3 levels of correction, so it is a fourth-order polynomial w.r.t. the weight vector:
\begin{equation}\label{eq:rig_function}
\begin{split}
f(\textbf{w}) = & \textbf{B}\textbf{w} + \sum_{i,j\in\mathcal{P}}w_iw_j\textbf{b}^{\{i,j\}} +  \sum_{i,j,k\in\mathcal{T}}w_iw_jw_k\textbf{b}^{\{i,j,k\}}+ \\
& \sum_{i,j,k,l\in\mathcal{Q}}w_iw_jw_kw_l\textbf{b}^{\{i,j,k,l\}}
\end{split}
\end{equation}
where $\textbf{B}\in\mathbb{R}^{3n\times m}$ is a blendshape matrix, $\mathcal{P,T,Q}$ are sets of tuples of size two, three and four respectively, containing controllers that need a corrective term and $\textbf{b}^{\{i,j\}},\textbf{b}^{\{i,j,k\}},\textbf{b}^{\{i,j,k,l\}}\in\mathbb{R}^{3n}$ are corrective blendshapes for each level of correction and specified tuple of controllers. 

\subsection{Approximation of the rig function} \label{proposed_modeling}

The rig function (\ref{eq:rig_function}) is a fourth-order polynomial, and this makes our objective function (\ref{eq:objective}) hard to work with. One possibility would be to train a machine learning model that considers a rig to be a black box. However, this demands a lot of data, and data acquisition is a costly and time-consuming task. We prefer a model-based approach; hence the alternative is approximating the rig function with a simpler form. 

\begin{figure}
    \centering
    \includegraphics[width=\linewidth]{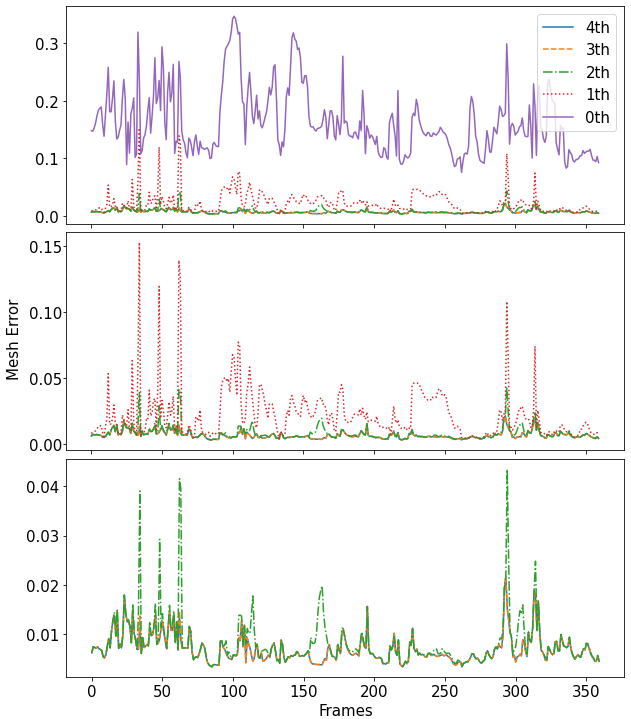}
    \caption{Mesh error (RMSE) over animation frames for different rig function approximations (see section \ref{proposed_modeling}). To give an idea of the error scale, we also included a 'zero approximation', i.e., in that case, the face is approximated by a neutral face at each time frame. $x$-axis represents frames of the animation, and $y$-axis the Euclidean offset of the meshes obtained via different rig approximations compared to the original animated  meshes.}
    \label{fig:approx_errors}
\end{figure}

A simple approximation can be provided by removing higher order corrective terms from the full rig function. The original rig in (\ref{eq:rig_function}) is quartic. A quadratic approximation is 
\begin{equation}\label{eq:q_approx}
    f(\textbf{w}) \approx \textbf{B}\textbf{w} + \sum_{i,j\in\mathcal{P}}w_iw_j\textbf{b}^{\{i,j\}}
\end{equation}
The simplest approximation is a linear one
\begin{equation}\label{eq:l_approx}
    f(\textbf{w}) \approx \textbf{B}\textbf{w}.
\end{equation}
We can see the error of each level approximation of the rig, over one of our datasets (\textit{DS 1}, see Section \ref{exper_setup}) in Figure \ref{fig:approx_errors}. Linear approximation produces significantly larger error compared to higher-order rig functions, while quadratic is quite comparable with cubic and quartic except in the frames where the error produces sudden peaks\footnote{Notice that there is a positive error even with a fourth level polynomial rig function. The reason is that in the animation process, additional corrections and refinements are added so that our facial model is not perfectly blendshape based. However, these errors are reasonably low, and we can consider that the fourth-order polynomial is a good enough approximation of the truth.}.

To solve our optimization problem (\ref{eq:objective}), we can work with the complexity level of linear and quadratic rig functions\footnote{Higher order corrections make the problem too complex.}. If linear approximation was able to  produce good estimates in inverse rig fitting, it would be a better choice due to its simplicity. In that case, we have a convex quadratic objective with box constraints:
\begin{equation}\label{eq:linear_obj}
\minimize_{0\leq\textbf{w}\leq1}\| \textbf{Bw} - \hat{\textbf{b}}\|^2 + \lambda\textbf{1}^T\textbf{w}
\end{equation}
that can be solved efficiently \cite{boyd2004convex}. In our experiments we will use a Python library \texttt{CVXPY} \cite{diamond2016cvxpy} to solve this problem. Due to its simplicity, a linear approximation (\ref{eq:l_approx}) is still used in the industry, and also corresponds to all model-based inverse rig solutions in the literature \cite{choe2001analysis, li2010example, yu2014regression, ccetinaslan2016position}, but we need to work with a more accurate quadratic approximation. The following section introduces the solution for solving the objective under a quadratic approximation (\ref{eq:q_approx}), which is our proposed algorithm.

\subsection{Proposed Algorithm}\label{first_quadratic}

The optimization problem (\ref{eq:objective}) under the quadratic rig approximation becomes
\begin{equation}\label{eq:qadratic_obj}
\minimize_{0\leq\textbf{w}\leq1}\|\textbf{Bw}+ \sum_{j,k\in\mathcal{P}}w_jw_k\textbf{b}^{\{j,k\}} - \hat{\textbf{b}}\|^2 + \lambda\textbf{1}^T\textbf{w}
\end{equation}
We approach this problem in a manner of LM, where at each iteration we have a vector of controllers' weights $\textbf{w}\in\mathbb{R}^m$ and need to solve for an optimal increment vector $\textbf{v}\in\mathbb{R}^m$ (\ref{eq:lm_iter}). Even under the quadratic approximation of the rig function, the objective function of LM iteration is fairly complex, hence we simplify the objective by applying MM. That is, to solve (\ref{eq:qadratic_obj}), we propose an LM-MM based method that is presented in Algorithm \ref{alg:cap}. We first explain the algorithm derivation and then detail each step of the algorithm.

The objective (\ref{eq:qadratic_obj}) consists of the data fidelity term $\|f(\textbf{w})-\hat{\textbf{b}}\|^2$ and the regularization term $\lambda\textbf{1}^T\textbf{w}$. If we write down the fidelity term as a sum, and consider each element $i$ of the sum separately, we can introduce a simpler notation. Namely, we introduce a symmetric (and sparse) matrix $\textbf{D}^{(i)}\in\mathbb{R}^{m\times m}$ for each face coordinate $i$. Nonzero entries of the matrix are extracted from the corrective blendshapes $D^{(i)}_{jk} = D^{(i)}_{kj} = \frac{1}{2}b_i^{\{j,k\}}$. With this we can represent the fidelity term in a canonical quadratic form: 
\begin{equation}
\begin{split}
\sum_{i=1}^{n}(\textbf{B}_i\textbf{w} + & \sum_{j,k\in\mathcal{P}}w_jw_kb_i^{\{j,k\}}-\hat{{b}}_i)^2  = 
\\ &  \sum_{i=1}^{n}(\textbf{B}_i\textbf{w}+ \textbf{w}^T\textbf{D}^{(i)}\textbf{w} -\hat{{b}}_i)^2
\end{split}
\end{equation}
Introduce function $\phi_i(\textbf{w}): \mathbb{R}^m\rightarrow\mathbb{R}$ as:
\begin{equation}
\phi_i(\textbf{w}) : = (\textbf{B}_i\textbf{w}+\textbf{w}^T\textbf{D}^{(i)}\textbf{w}-\hat{{b}}_i)^2
\end{equation}
When we add the increment vector $\textbf{v}$ on top of the current weight vector $\textbf{w}$ it yields:
\begin{equation}
\begin{split}
\phi_i(\textbf{w}+\textbf{v}) & = (g_i + \textbf{h}_i\textbf{v} + \textbf{v}^T\textbf{D}^{(i)}\textbf{v})^2 \\
& = g_i^2 + 2g_i\textbf{h}_i\textbf{v} + 2g_i\textbf{v}^T\textbf{D}^{(i)}\textbf{v} + (\textbf{h}_i\textbf{v} + \textbf{v}^T\textbf{D}^{(i)}\textbf{v})^2
\end{split}
\end{equation}
where $g_i:=\textbf{B}_i\textbf{w} + \textbf{w}^T\textbf{D}^{(i)}\textbf{w} - \hat{b}_i$, and $\textbf{h}_i := \textbf{B}_i + 2\textbf{w}^T\textbf{D}^{(i)}$ are introduced to simplify the notation. The fidelity term from (\ref{eq:qadratic_obj}) is a sum of functions $\phi_i(\textbf{w})$, hence in order to bound the objective, we will derive an upper bound $\psi_i(\textbf{v};\textbf{w})\geq\phi_i(\textbf{w}+\textbf{v})$ for each element of the sum (see Section \ref{preliminaries_mm}, (\ref{eq:majorizer})). Functions $\psi_i(\textbf{v};\textbf{w})$ depend only on $\textbf{v}$, while $\textbf{w}$ is considered fixed, so we will drop it in the equations. Let us first separate $\phi_i(\textbf{w})$ into $x_i := 2g_i\textbf{v}^T\textbf{D}^{(i)}\textbf{v}$ and $y_i:=(\textbf{h}_i\textbf{v} + \textbf{v}^T\textbf{D}^{(i)}\textbf{v})^2$, and bound each term separately. Bound on $x_i$ depends on the sign of $g_i$, so if we define a function
$$  \lambda_M(\textbf{D}^{(i)},g_i) :=
    \begin{cases}
      \lambda_{\text{min}}(\textbf{D}^{(i)}) & \text{if } g_i<0\\
      \lambda_{\text{max}}(\textbf{D}^{(i)}) & \text{if } g_i\geq0
    \end{cases} 
$$
we can write the bound as 
$$2g_i\textbf{v}^T\textbf{D}^{(i)}\textbf{v} \leq 2g_i\lambda_M(\textbf{D}^{(i)},g_i)\|\textbf{v}\|^2.$$
The bound on $y_i$ is obtained by applying the Cauchy-Schwartz inequality multiple times:
\begin{equation}
\begin{split}
(\textbf{h}_i\textbf{v} + \textbf{v}^T\textbf{D}^{(i)}\textbf{v})^2 & \leq 2(\textbf{h}_i\textbf{v})^2 + 2(\textbf{v}^T\textbf{D}^{(i)}\textbf{v})^2\\
& \leq  2\|\textbf{h}_i\|^2\|\textbf{v}\|^2 + 2 \|\textbf{v}\|^4\|\textbf{D}^{(i)}\|^2 \\
& \leq 2\|\textbf{h}_i\|^2\|\textbf{v}\|^2 + 2m\sigma^2_{\text{max}}(\textbf{D}^{(i)})\sum_{j=1}^mv_j^4.
\end{split}
\end{equation}

The bound function $\psi_i(\textbf{v})$ for a coordinate $i$ is then:
\begin{equation}
\begin{split}
\psi_i(\textbf{v}) := & g_i^2 + 2g_i\sum_{j=1}^m h_{ij}v_j + \\  
2(g_i\lambda_M( &\textbf{D}^{(i)},g_i) + \|\textbf{h}_i\|^2)\sum_{j=1}^mv_j^2 + 2m\sigma_{\text{max}}^2(\textbf{D}^{(i)})\sum_{j=1}^mv_j^4
\end{split}
\end{equation}
and the bound for the complete fidelity term is the sum of coordinate-wise bounds:
\begin{equation}\label{eq:bound_psi}
    \psi(\textbf{v}) = \sum_{i=1}^n \psi_i(\textbf{v}).
\end{equation}
The problem to be solved at each MM iteration $t$, with $\textbf{w}=\textbf{w}_{(t)}$, is then:
\begin{equation}\label{eq:derivation_lm_tr}
    \begin{split}
        \minimize_{\textbf{v}} \,\, & \psi(\textbf{v};\textbf{w}) + \lambda\textbf{1}^T(\textbf{w} + \textbf{v}) \\
        \text{s.t.   }  & \textbf{0}\leq \textbf{w}+\textbf{v} \leq \textbf{1}
    \end{split}
\end{equation}
Finally, observe that this problem can be solved for each controller $j=1,...,m$ separately, where the per-component problem has a quartic one-dimensional form with the cubic coefficient equal to zero:
\begin{equation}\label{eq:derivation_component}
    \begin{split}
        \minimize_{v_j} \,\, & p + qv_j + rv_j^2 + sv_j^4 \\
        \text{s.t.   } &  0\leq w_j+v_j \leq 1
    \end{split}
\end{equation}
and the coefficients $p$, $q$, $r$, and $s$ are:
\begin{equation}\label{eq:coefficients}
\begin{split}
p:= & \sum_{i=1}^n g_i^2 + \lambda\textbf{1}^T\textbf{w} = \|\textbf{B}\textbf{w} + \sum_{i=1}^n\textbf{w}^T\textbf{D}^{(i)}\textbf{w} - \hat{\textbf{b}}\|^2 + \lambda\textbf{1}^T\textbf{w} \\
q:= & 2\sum_{i=1}^m g_ih_{ij} + \lambda \\
  = & 2(\textbf{Bw}+\sum_{i=1}^n\textbf{w}^T\textbf{D}^{(i)}\textbf{w}-\hat{\textbf{b}})^T(\textbf{B}_{:j} + 2\textbf{w}^T\textbf{D}^{(i)}_j) + \lambda \\
r:= & 2\sum_{i=1}^n(g_i\lambda_M(\textbf{D}^{(i)},g_i)+\|\textbf{h}_i\|^2) = \\
 2 \sum_{i=1}^n  & ((\textbf{B}_i \textbf{w} + \textbf{w}^T\textbf{D}^{(i)}\textbf{w}-\hat{b}_i)\lambda_M(\textbf{D}^{(i)},\textbf{B}_i\textbf{w} + \textbf{w}^T\textbf{D}^{(i)}\textbf{w} - \hat{b}_i) \\
 & + \|\textbf{B}_i+2\textbf{w}^T\textbf{D}^{(i)} \|^2) \\
s:= & 2m\sum_{i=1}^n\sigma_{\text{max}}^2(\textbf{D}^{(i)})
\end{split}
\end{equation}

Notice that the coefficient $q$ depends on a coordinate $j$, so it has to be computed for each controller separately, while $p$, $r$, and $s$ are computed only once per iteration. We can find the extreme values of the polynomial using the roots of the cubic derivative and check if they are within the feasible interval, and also compare with the polynomial values at the borders, to get the constrained minimizer of problem (\ref{eq:derivation_component}).

Summarizing, our solver of (\ref{eq:qadratic_obj}) generates an iterate sequence $\textbf{w}_{(t)}$ as follows.
 At each iteration $t$, we seek for an increment $\textbf{v}$ such that the next iterate is set to $\textbf{w}_{(t+1)}=\textbf{w}_{(t)}+\textbf{v}$. The increment $\textbf{v}$ is sought via an MM method by solving (\ref{eq:derivation_lm_tr}). The solution of (\ref{eq:derivation_lm_tr}) is in turn obtained by solving problems (\ref{eq:derivation_component}) component-wise.

The pseudocode for solving the inverse rig under a proposed upper bound is presented in Algorithm \ref{alg:cap}. Notice that in one of the steps of the algorithm we compute eigen and singular values of matrices $D^{(i)}$. This computation is needed only once per character, and we can reuse the computed values for each following frame that is to be fitted. 
The algorithm terminates either if it reaches a specified maximal number of iterations $T$ or if the cost for iterate $\textbf{w}_{(t)}$:
\begin{equation}\label{residuals}
    g(\textbf{w}_{(t)}) = \|f(\textbf{w}_{(t)}) - \hat{\textbf{b}}\|^2 + \lambda\textbf{1}^T\textbf{w}_{(t)}
\end{equation}
does not change between two consecutive iterations more than a specified tolerance $\epsilon>0$. The algorithm can be initialized in principle by any $\textbf{w} \in [0,1]^m$. To obtain faster convergence, we use the initialization schemes detailed in Section \ref{exper_setup}.

\begin{algorithm}
\caption{}\label{alg:cap}
\begin{algorithmic}
\Require $\lambda>0$, $\epsilon>0$, $\hat{\textbf{b}}\in\mathbb{R}^{3n}$, $\textbf{B}\in\mathbb{R}^{3n\times m}$, $D^{(i)}\in\mathbb{R}^{m\times m}$ for $i=1,...,3n$, $\textbf{w}_{(0)}\in[0,1]^m$, $T\in\mathbb{N}$.
\Ensure $\hat{\textbf{w}}$ - an approximate minimizer of the problem (\ref{eq:qadratic_obj}).
\State Compute singular and eigen values $\lambda_{\text{min}}(D^{(i)})$, $\lambda_{\text{max}}(D^{(i)})$, $\sigma_{\text{max}}(\textbf{D}^{(i)})$ for $i=1,...,3n$.
\For{$t = 1,...,T$}
\State Compute $g(\textbf{w}_{(t)})$ using (\ref{residuals})
\State Check convergence:
\If{$t > 1$}
    \If{$|g(\textbf{w}_{(t-1)})|<\epsilon$}
        \State \Return \textbf{w}
    \EndIf
\EndIf
\State Compute coefficients $p,r$ and $s$ using (\ref{eq:coefficients}):
\State $p=  \sum_{i=1}^{3n} g_i^2 + \lambda\textbf{1}^T\textbf{w}$
\State $r=  2\sum_{i=1}^{3n}(g_i\lambda_M(\textbf{D}^{(i)},g_i)+\|\textbf{h}_i\|^2)$
\State $s=  2m\sum_{i=1}^{3n}\sigma_{\text{max}}^2(\textbf{D}^{(i)})$
\For{$j=1,...,m$} 
    \State Compute $q=q_j$ using (\ref{eq:coefficients}):
    \State $q=  2\sum_{i=1}^m g_ih_{ij} + \lambda$
    \State Solve for the optimal increment $\hat{v}_j$:
    \State $\hat{v}_j = \argmin_{v} p + qv + rv^2 + sv^4 $
    \State \qquad s.t. $-w^{(t)}_j\leq v\leq 1-w^{(t)}_j$
\EndFor
\State Update the iterate $\textbf{w}_{(t)}$:
\State $\textbf{w}_{(t+1)} = \textbf{w}_{(t)} + \hat{\textbf{v}}$
\EndFor
\State \Return \textbf{w}
\end{algorithmic}
\end{algorithm}
By the standard MM theory \cite{becker1997algorithms, lange2000optimization, hunter2004tutorial, zhang2007surrogate} and construction of Algorithm \ref{alg:cap}, 
the estimate sequence $\textbf{w}_{(t)}$ is feasible to problem (\ref{eq:qadratic_obj}) at all iterations $t$, and moreover the sequence of values of the objective in (\ref{eq:qadratic_obj}) evaluated at $\textbf{w}_{(t)}$ is non-increasing. 
As demonstrated numerically in Section \ref{subsec_num_res}, Algorithm \ref{alg:cap} produces accurate and sparse solutions of the inverse rig problem.


\section{Results}\label{section_results}

In this section we show and discuss numerical results over several animation datasets. Section \ref{exper_setup} first introduces the data used for evaluation as well as the experimental setup. Section \ref{subsec_num_res} then shows the obtained numerical results over each of the datasets.

\subsection{Experimental Setup}\label{exper_setup} 
\begin{figure}
    \centering
    \includegraphics[width=.7\linewidth]{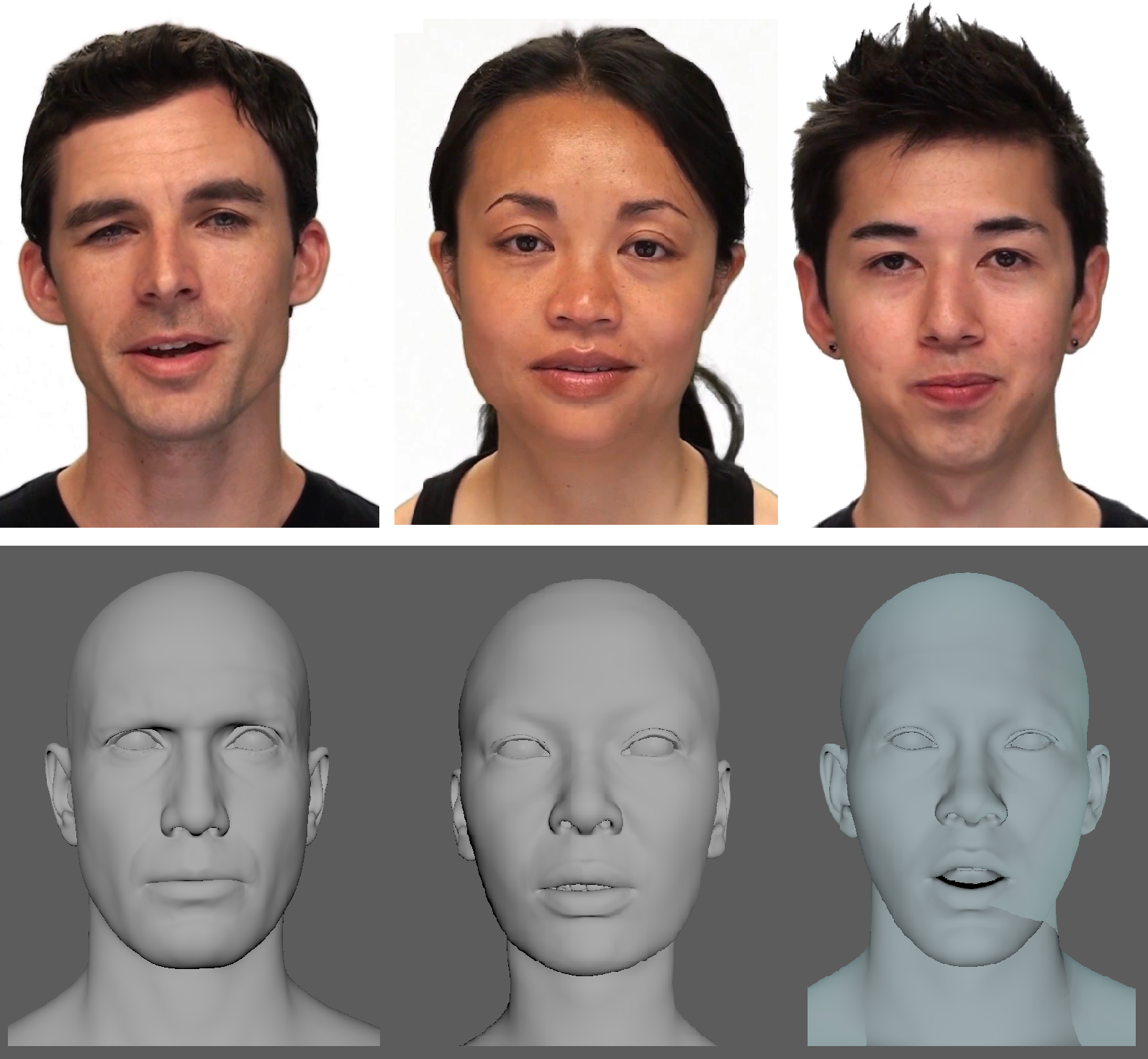}
    \caption{In the upper row of a figure we see three RAVDSS actors, and in the bottom row, three \texttt{MetaHumans} characters, animated with the corresponding actor.}
    \label{fig:actors}
\end{figure}

Five animation datasets were used here to evaluate the proposed algorithm. The first two datasets --- \textit{DS 1} and \textit{DS 2} are proprietary animations, provided to us by the 3Lateral Studio\footnote{https://www.3lateral.com/} for the purposes of the paper. We collected additional three datasets (\textit{DS 3 --- DS 5}) that are publicly available so that the reader has the possibility of recreating the experiments. These latter sets are created using the \texttt{MetaHumans} software\footnote{https://www.unrealengine.com/en-US/digital-humans} in combination with RAVDESS emotional speech performance \cite{livingstone_steven_r_2018_1188976}. \texttt{MetaHumans Creator} provides many free characters for animation, and it allows a non-expert to animate any of them by connecting it to a captured video of the facial performance. We used 3 RAVDESS actor --- \texttt{MetaHumans} character pairs: Actor 1 --- Omar, Actor 18 --- Danielle and Actor 21 --- Myles (These pairs are depicted in Figure \ref{fig:actors}). 

To facilitate reproducibility, we created an online directory (\url{https://zenodo.org/record/5329677#.YSvQ9I4zaCo}) that contains datasets \textit{DS 3---DS 5} i.e., extracted meshes, weights, and a blendshape basis in forms of \texttt{numpy} arrays. It also contains \texttt{Python} scripts used to extract data and run Algorithm \ref{alg:cap} over the animation and also a \texttt{CVXPY}-based solver of the linear model (\ref{eq:linear_obj})\footnote{The scripts with the proposed algorithm are not present in the directory at the moment, and will be added upon acceptance of the paper.}. 

In this section, we compare the performance of Algorithm \ref{alg:cap}, which uses a quadratic approximation of the rig, and a standard model that considers a linear approximation and solves eq. (\ref{eq:linear_obj}). For the latter we use the \texttt{Python} library \texttt{CVXPY}\footnote{https://www.cvxpy.org/}, that solves (constrained) convex problems.

We next describe the initialization of Algorithm \ref{alg:cap}. As mentioned earlier, we look for a sparse solution, so a zero vector $\textbf{w}_{(0)}=\textbf{0}$ is a good choice (we refer to this choice as the \textit{zero} initialization). Another approach is using an approximate guess of the solution. As discussed in \cite{ccetinaslan2016position}, in the case of a linear rig function ($f(\textbf{w})=\textbf{Bw}$) without any constraints, a least-squares solution is $\hat{\textbf{w}} = \textbf{B}^{\dagger}\hat{\textbf{b}}$ where $\hat{\textbf{b}}$ is a given mesh. As this is a fairly easy value to obtain (pseudoinverse is performed only once per character and then reused for different frames in animation), we consider it as another choice of the initial point. However, we have to respect the constraints of our model, so we project this vector to be within 0-1 range, hence the initial point in this case is $\textbf{w}_{(0)} = P_{[0,1]}(\textbf{B}^{\dagger}\hat{\textbf{b}})$ (we refer to this choice as the \textit{pseudoinverse} initialization). Finally, as we expect (and it will show to be true) that a linear model obtained via \texttt{CVXPY} is not as accurate, but it is faster to compute, we can use this linear solution as yet another initialization strategy. That is, we refer to a solution of (\ref{eq:linear_obj}) as the \textit{linear} initialization. Note that, as (\ref{eq:qadratic_obj}) is a nonconvex problem, the final solution obtained may depend on the adopted initialization.

We compare the proposed Algorithm \ref{alg:cap} with a numerical solver (see Section \ref{proposed_modeling} for details) of the linear rig-based formulation (\ref{eq:linear_obj}). The latter is a solid representative of state of the art for model-based inverse rig solutions \cite{choe2001analysis, li2010example, yu2014regression, ccetinaslan2016position} that always utilize the inverse rig approximation.

Dimensions differ over the datasets, but the order of magnitude is similar. Number of vertices for \textit{DS 1} and \textit{DS 2} is $n=5863$, and for the other datasets it is $n=6012$. Number of controllers $m$ ranges between 62 and 147, while the number of frames used in the experiments $N$ takes values from 142 to 204. A more detailed description of our data can be found at \url{https://zenodo.org/record/5329677#.YSvQ9I4zaCo}.

\subsection{Numerical Results}\label{subsec_num_res} 

The output of the inverse rig problem is a vector of weights $\hat{\textbf{w}}$, and once we have it, the animation software\footnote{https://www.autodesk.com/products/maya/overview} uses a rig function $f(\hat{\textbf{w}})$ to give the resulting face mesh. The principal metric of interest is the mesh error, which is computed as a root mean squared error (RMSE) between this estimate and a given mesh $\hat{\textbf{b}}$:
$$\frac{\|f(\hat{\textbf{w}})-\hat{\textbf{b}}\|}{n}.$$ This we call a \textit{fidelity measure} of the solution. Here we do not use the function approximations but a "true" rig function with all the corrective terms --- this will give a fair comparison between methods that use linear (\ref{eq:l_approx}) and quadratic (\ref{eq:q_approx}) approximation.

As mentioned earlier in the paper, we are also interested in a \textit{sparsity measure}. For this, we use cardinality of the estimated vector $\hat{\textbf{w}}$. Finally, we also include the number of iterations each algorithm takes to solve the problem.

\begin{figure}
    \centering
    \includegraphics[width=\linewidth]{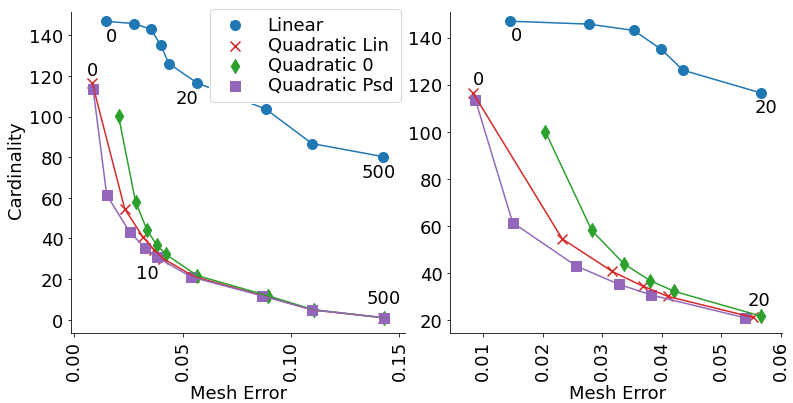}
    \caption{Trade-off between Mesh Error and Cardinality of the estimated solution for \textit{DS 1} (averaged over the frames). Points in the scatter correspond to different values of the regularization parameter $\lambda$. Left subfigure covers a complete (mentioned) range of $\lambda$ values, while the right subfigure zoom-in to better notice the differences between the curves. We observe linear model (\texttt{CVXPY} solution) and a quadratic with three initialization approaches (\textit{linear}, \textit{zero} and \textit{pseudoinverse}).}
    \label{fig:hm_tradeoff}
\end{figure}
\begin{figure}
    \centering
    \includegraphics[width=\linewidth]{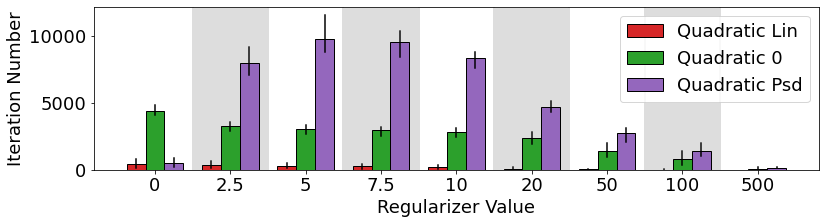}
    \caption{Iteration number for \textit{DS 1} for three different initializations of the algorithm (\textit{linear}, \textit{zero} and \textit{pseudoinverse}). $x$-axis corresponds to different values of the regularization parameter $\lambda$; bar height shows the median number of iterations over all the frames, and a line on top of a bar indicates the upper and the lower quartiles.}
    \label{fig:hm_iter}
\end{figure}
\begin{figure}
    \centering
    \includegraphics[width=\linewidth]{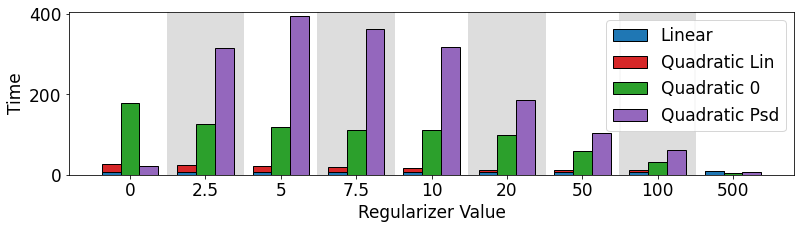}
    \caption{Average execution time for \textit{DS 1} for different approaches. $x$-axis corresponds to different values of the regularization parameter $\lambda$; bar height shows the average execution time over all the frames.}
    \label{fig:hm_time}
\end{figure}

Let us at the moment consider only the first dataset \textit{DS 1}. In the objective (\ref{eq:objective}) there is a parameter $\lambda\geq 0$ for enforcing regularization (sparsity). We run experiments with a range of values $\lambda\in\{0,2.5,5,7.5,10,20,50,100,500\}$ to observe the performances of algorithms. Figure \ref{fig:hm_tradeoff} shows a trade-off curve between cardinality and fidelity. A curve for linear model approximation is always above the others, indicating, as expected, inferior performances compared to the quadratic model. For the quadratic case, we see that the performances of three initializations are similar, exhibiting mutual differences that are smaller compared to their gains w.r.t. linear model-based solution in (\ref{eq:linear_obj}), with \textit{linear} initialization exhibiting slightly higher error. For all four curves, the elbow pattern appears for values of $\lambda\in\{10,20\}$, indicating that this is the regularization term that leads to the best trade-off between the accuracy and the sparsity of the solution.  An important difference between different initializations is the number of iterations needed for convergence, shown in Figure \ref{fig:hm_iter}. Each bar represents the average number of iterations over the frames and is accompanied by lines that indicate lower and upper quartile, and $x$-axis corresponds to the regularization parameter $\lambda$. The number of iterations is a good indicator of the execution time for the algorithm, which is confirmed by the similarity of Figures \ref{fig:hm_iter} and \ref{fig:hm_time}. In Figure \ref{fig:hm_time} green bar is stacked on top of a blue one, to have more realistic representation of the execution time, since the \textit{linear} initialization should also account for the time needed to obtain a solution of the linear approach (\ref{eq:linear_obj}).
Except in the case with $\lambda=0$ (no regularization term), the \textit{pseudoinverse} initialization approach is far slower than the other two. Since it produces results similar to the zero initialization, we will dismiss it as a more demanding alternative. On the other hand, \textit{linear} initialization leads to fast convergence, often demanding just a few iterations in addition to a fast-to-compute initial vector.

\begin{figure}
    \centering
    \includegraphics[width=\linewidth]{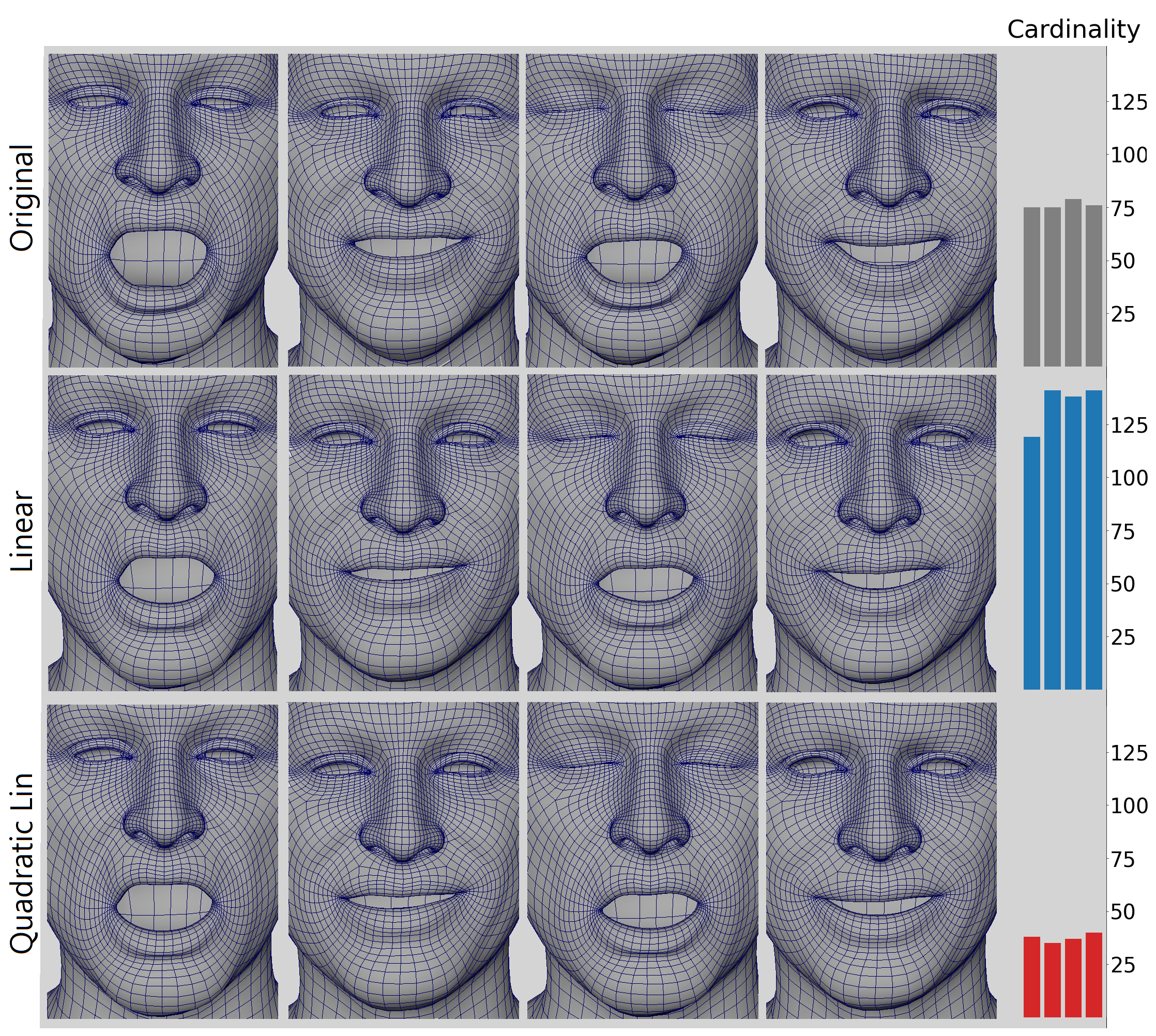}
    \caption{Mesh details corresponding to four frames with the highest (average) mesh error for \textit{DS1} and $\lambda=10$. The upper row shows the original (ground-truth) meshes; the middle row represents the estimated meshes using the linear rig approximation; the bottom row depicts the results of our model. In the right-most column of the figure, we show the average cardinality of the frames.}
    \label{fig:reconstructions}
\end{figure}

In Figure \ref{fig:reconstructions}, we present the estimated meshes corresponding to the frames with the highest fidelity error. Visual inspection tells us that the most apparent misalignment with the target is in the mouth region. The improvements of the quadratic model are not easy to spot here (pay attention to the cheeks and the lower lip). Still, the cardinality is vastly reduced and even considerably lower compared to the original frames. 

\begin{figure}
    \centering
    \includegraphics[width=\linewidth]{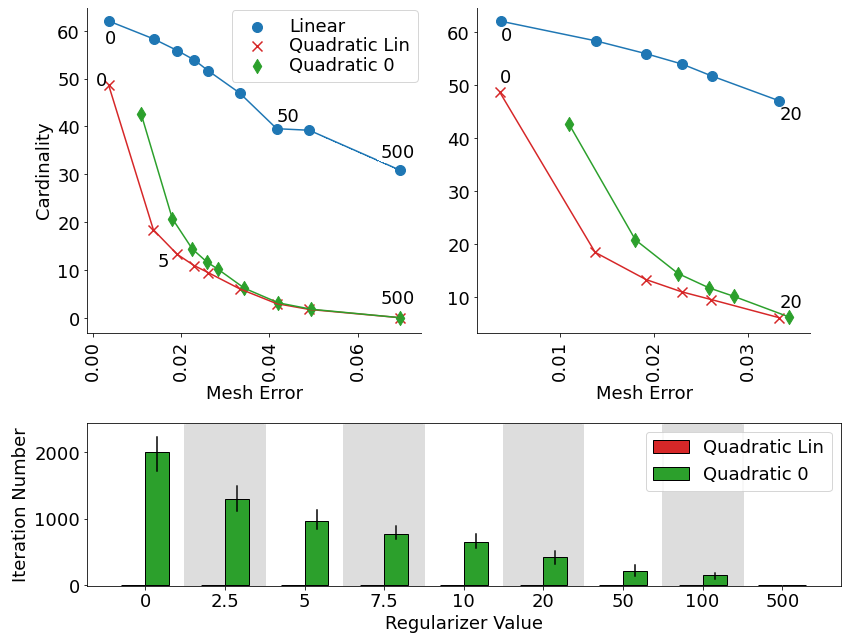}
    \caption{DS 2. Upper row: Trade-off between Mesh Error and Cardinality of the estimated solution. Bottom row: Number of iterations to convergence, for varying values of $\lambda$.}
    \label{fig:o_metrics}
\end{figure}

\begin{figure}
    \centering
    \includegraphics[width=\linewidth]{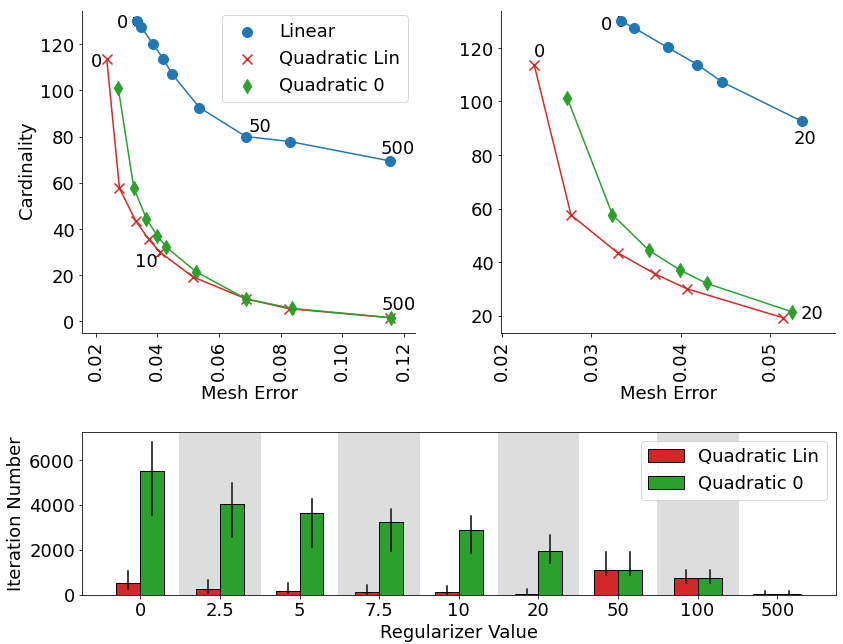}
    \caption{DS 3. Upper row: Trade-off between Mesh Error and Cardinality of the estimated solution. Bottom row: Number of iterations to convergence, for varying values of $\lambda$.}
    \label{fig:a1_metrics}
\end{figure}

\begin{figure}
    \centering
    \includegraphics[width=\linewidth]{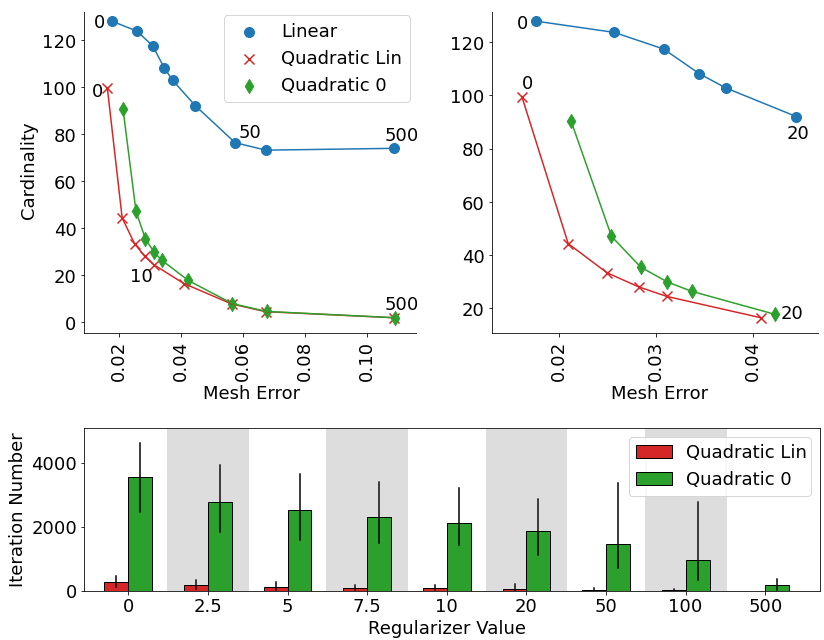}
    \caption{DS 4. Upper row: Trade-off between Mesh Error and Cardinality of the estimated solution. Bottom row: Number of iterations to convergence, for varying values of $\lambda$.}
    \label{fig:a18_metrics}
\end{figure}

\begin{figure}
    \centering
    \includegraphics[width=\linewidth]{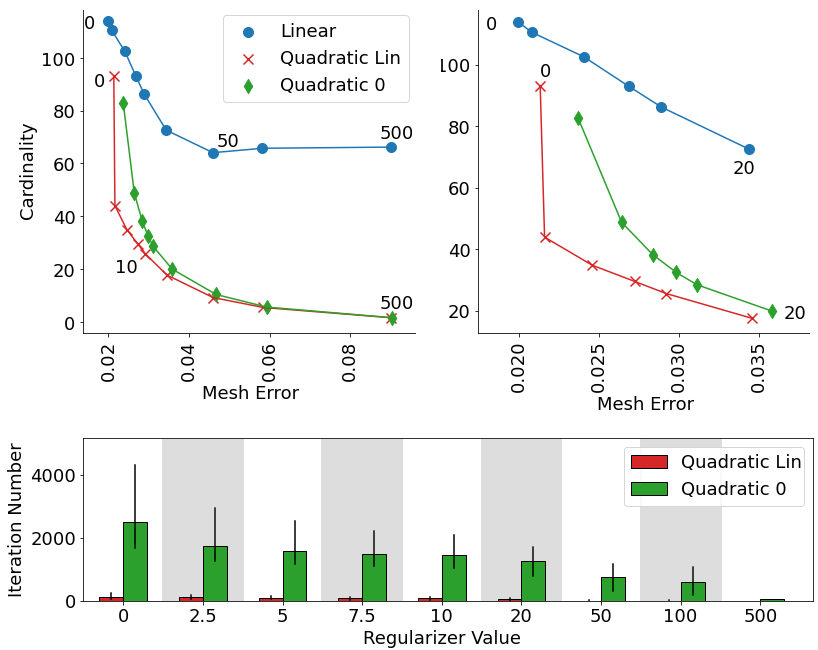}
    \caption{DS 5. Upper row: Trade-off between Mesh Error and Cardinality of the estimated solution. Bottom row: Number of iterations to convergence, for varying values of $\lambda$.}
    \label{fig:a21_metrics}
\end{figure}

For datasets \textit{DS 2 --- DS 5} we exclude the \textit{pseudoinverse} initialization, and the results are presented in figures \ref{fig:o_metrics}-\ref{fig:a21_metrics}. The behavior of trade-off curves and iteration numbers in all the cases is similar to those of \textit{DS 1}. Interestingly, in all of these datasets, the elbow pattern for a linear model appears to be for $\lambda=50$ (higher than for the \textit{DS 1}), while the shape of the curves corresponding to the quadratic approximation does not change as much.
\textit{DS 3 --- DS 5} have a larger number of vertices, and hence also demand more iterations compared to \textit{DS 2}. Still, for \textit{DS 2} the difference between the number of iterations in two initialization approaches is even more significant, where \textit{linear} initialization finds a solution almost immediately. Based on this, and the fact that the trade-off curve for \textit{linear} initialization is always below the other two, we might conclude that this approach --- combining the convex inverse rig solution (\ref{eq:linear_obj}) with Algorithm \ref{alg:cap} --- is the winning solution. 

\section{Conclusion}\label{section_conclusion}

In this paper we proposed a novel, fast, sparse, and accurate algorithm for solving the inverse rig problem in facial blendshape animation. The algorithm is model-based i.e., it takes into account the (approximate) rig function as opposed to the data-based approaches that demand huge amounts of data for training. In contrast with other proposed model-based solutions, we consider a quadratic instead of the linear rig function approximation. This increases the complexity of the problem, but also increases the accuracy of the solution. Capitalizing on LM method, MM, and sparsity-promoting regularization, we develop an efficient solver for the corresponding nonconvex constrained least squares problem. The proposed method makes the major computational step therein -- minimizing a MM-like surrogate function at each iteration -- decoupled across components of the vector of controller weights $\textbf{w}$ and is hence amenable for parallelization.

We carried out extensive experiments on a number of open and proprietary animated characters data sets. The results show that our algorithm exhibits not only the higher data fidelity of the predicted meshes compared to a standard linear model, but also the predicted vector of weights has considerably fewer activated components, which is a desirable treat of the solution. At the same time, the proposed method incurs a small computational overhead with respect to the linear rig-based solution.


\begin{thebibliography}{1}

\bibitem{mori2012uncanny}
Mori, Masahiro, Karl F. MacDorman, and Norri Kageki. "The uncanny valley [from the field]." IEEE Robotics \& Automation Magazine 19.2 (2012): 98-100.

\bibitem{choe2001analysis}
Choe, Byoungwon, and Hyeong-Seok Ko. "Analysis and synthesis of facial expressions with hand-generated muscle actuation basis." ACM SIGGRAPH 2006 Courses. 2006. 21-es.

\bibitem{li2010example}
Li, Hao, Thibaut Weise, and Mark Pauly. "Example-based facial rigging." ACM Transactions on Graphics (TOG) 29.4 (2010): 1-6.

\bibitem{joshi2006learning}
Joshi, Pushkar, et al. "Learning controls for blend shape based realistic facial animation." ACM Siggraph 2006 Courses. 2006. 17-es.

\bibitem{yu2014regression}
Yu, Hui, and Honghai Liu. "Regression-based facial expression optimization." IEEE Transactions on Human-Machine Systems 44.3 (2014): 386-394.

\bibitem{ccetinaslan2016position}
Cetinaslan, Cumhur Ozan. "Position Manipulation Techniques for Facial Animation." Faculdade de Ciencias da Universidade do Porto (2016).

\bibitem{holden2015learning}
Holden, Daniel, Jun Saito, and Taku Komura. "Learning an inverse rig mapping for character animation." Proceedings of the 14th ACM SIGGRAPH/Eurographics Symposium on Computer Animation. 2015.

\bibitem{holden2016learning}
Holden, Daniel, Jun Saito, and Taku Komura. "Learning Inverse Rig Mappings by Nonlinear Regression." IEEE Transactions on Visualization and Computer Graphics 23.3 (2016): 1167-1178.

\bibitem{boyd_2011}
Boyd, Stephen, Neal Parikh, and Eric Chu. Distributed optimization and statistical learning via the alternating direction method of multipliers. Now Publishers Inc, 2011.

\bibitem{cao_2017}
Cao, Bin, et al. "Distributed parallel particle swarm optimization for multi-objective and many-objective large-scale optimization." IEEE Access 5 (2017): 8214-8221.

\bibitem{Notarnicola2018}
Notarnicola, Ivano, Ruggero Carli, and Giuseppe Notarstefano. "Distributed partitioned big-data optimization via asynchronous dual decomposition." IEEE Transactions on Control of Network Systems 5.4 (2017): 1910-1919.

\bibitem{jakovetic2020primal}
Jakovetić, Dušan, et al. "Primal–dual methods for large-scale and distributed convex optimization and data analytics." Proceedings of the IEEE 108.11 (2020): 1923-1938.

\bibitem{lewis2014practice}
Lewis, John P., et al. "Practice and Theory of Blendshape Facial Models." Eurographics (State of the Art Reports) 1.8 (2014): 2.

\bibitem{alkawaz2015blend}
Alkawaz, Mohammed Hazim, et al. "Blend shape interpolation and {FACS} for realistic avatar." 3D Research 6.1 (2015): 6.

\bibitem{boyd2004convex}
Stephen P. Boyd, and Lieven Vandenberghe. Convex optimization. Cambridge university press, 2004.

\bibitem{Levenberg1944AMF}
Levenberg, Kenneth. "A method for the solution of certain non-linear problems in least squares." Quarterly of Applied Mathematics 2.2 (1944): 164-168.

\bibitem{marquardt1963algorithm}
Marquardt, Donald W. "An algorithm for least-squares estimation of nonlinear parameters." Journal of the society for Industrial and Applied Mathematics 11.2 (1963): 431-441.

\bibitem{becker1997algorithms}
Becker, Mark P., Ilsoon Yang, and Kenneth Lange. "EM algorithms without missing data." Statistical Methods in Medical Research 6.1 (1997): 38-54.

\bibitem{lange2000optimization}
Lange, Kenneth, David R. Hunter, and Ilsoon Yang. "Optimization transfer using surrogate objective functions." Journal of Computational and Graphical Statistics 9.1 (2000): 1-20.

\bibitem{hunter2004tutorial}
Hunter, David R., and Kenneth Lange. "A tutorial on MM algorithms." The American Statistician 58.1 (2004): 30-37.

\bibitem{zhang2007surrogate}
Zhang, Zhihua, James T. Kwok, and Dit-Yan Yeung. "Surrogate maximization/minimization algorithms and extensions." Machine Learning 69.1 (2007): 1-33.

\bibitem{kelley1999iterative}
Kelley, Carl T. Iterative methods for optimization. Society for Industrial and Applied Mathematics, 1999.

\bibitem{yuan2000review}
Yuan, Ya-xiang. "A review of trust region algorithms for optimization." ICIAM Vol. 99. No. 1. 2000.

\bibitem{ranganathan2004levenberg}
Ranganathan, Ananth. "The Levenberg-Marquardt algorithm." Honda Research Institute USA (2012)

\bibitem{pujol2007solution}
Pujol, Jose. "The solution of nonlinear inverse problems and the Levenberg-Marquardt method." Geophysics 72.4 (2007): W1-W16.

\bibitem{lewis2010direct}
Lewis, John P., and Ken-Ichi Anjyo. "Direct manipulation blendshapes." IEEE Computer Graphics and Applications 30.4 (2010): 42-50.

\bibitem{li2013realtime}
Li, Hao, et al. "Realtime facial animation with on-the-fly correctives." ACM Transactions on Graphics 32.4 (2013): 42-1.

\bibitem{neumann2013sparse}
Neumann, Thomas, et al. "Sparse localized deformation components." ACM Transactions on Graphics (TOG) 32.6 (2013): 1-10.

\bibitem{song2017sparse}
Song, Jaewon, et al. "Sparse Rig Parameter Optimization for Character Animation." Computer Graphics Forum. Vol. 36. No. 2. 2017.

\bibitem{na2011local}
Na, Kyung-Gun, and Moon-Ryul Jung. "Local shape blending using coherent weighted regions." The Visual Computer 27.6 (2011): 575-584.


\bibitem{tena2011interactive}
Tena, J. Rafael, Fernando De la Torre, and Iain Matthews. "Interactive region-based linear 3d face models." ACM SIGGRAPH 2011 papers. 2011. 1-10.

\bibitem{fratarcangeli39fast}
Fratarcangeli, Marco, et al. "Fast Nonlinear Least Squares Optimization of Large‐Scale Semi‐Sparse Problems." Computer Graphics Forum. Vol. 39. No. 2. 2020.

\bibitem{romeo2020data}
Romeo, Marco, and Sara C. Schvartzman. "Data‐Driven Facial Simulation." Computer Graphics Forum. Vol. 39. No. 6. 2020.

\bibitem{rackovic}
Rackovic, Stevo, Soares, Claudia, Jakovetic, Dusan, Desnica, Zoranka and Ljubobratovic, Relja. "Clustering of the Blendshape Facial Model." To appear in Proceedings of EUSIPCO. 2021.

\bibitem{sifakis2005automatic}
Sifakis, Eftychios, Igor Neverov, and Ronald Fedkiw. "Automatic determination of facial muscle activations from sparse motion capture marker data." ACM SIGGRAPH 2005 Papers. 2005. 417-425.

\bibitem{ruder2016overview}
Dogo, E. M., et al. "A comparative analysis of gradient descent-based optimization algorithms on convolutional neural networks." 2018 International Conference on Computational Techniques, Electronics and Mechanical Systems (CTEMS). IEEE, 2018.

\bibitem{wang2012gauss}
Wang, Yong. "Gauss–Newton method." Wiley Interdisciplinary Reviews: Computational Statistics 4.4 (2012): 415-420.

\bibitem{li2019convergence}
Li, Xiaoyu, and Francesco Orabona. "On the convergence of stochastic gradient descent with adaptive stepsizes." The 22nd International Conference on Artificial Intelligence and Statistics. PMLR, 2019.

\bibitem{7131516}
Pitaval, Renaud-Alexandre, Wei Dai, and Olav Tirkkonen. "Convergence of gradient descent for low-rank matrix approximation." IEEE Transactions on Information Theory 61.8 (2015): 4451-4457.

\bibitem{dennis1996numerical}
Dennis Jr, John E., and Robert B. Schnabel. Numerical methods for unconstrained optimization and nonlinear equations. Society for Industrial and Applied Mathematics, 1996.

\bibitem{siregar2018analysis}
Siregar, Rahmi Wahidah, and Marwan Ramli. "Analysis local convergence of Gauss-Newton method." IOP Conference Series: Materials Science and Engineering. Vol. 300. No. 1. IOP Publishing, 2018.

\bibitem{yamashita2001rate}
Yamashita, Nobuo, and Masao Fukushima. "On the rate of convergence of the Levenberg-Marquardt method." Topics in Numerical Analysis. Springer, Vienna, 2001. 239-249.

\bibitem{fan2005quadratic}
Fan, Jin-yan, and Ya-xiang Yuan. "On the quadratic convergence of the Levenberg-Marquardt method without nonsingularity assumption." Computing 74.1 (2005): 23-39.

\bibitem{ahookhosh2019local}
Ahookhosh, Masoud, et al. "Local convergence of the Levenberg–Marquardt method under Holder metric subregularity." Advances in Computational Mathematics 45.5 (2019): 2771-2806.

\bibitem{yuan1994trust}
Yuan, Ya-xiang. "Trust region algorithms for nonlinear equations." Hong Kong Baptist University, Department of Mathematics, 1994.

\bibitem{conn2000trust}
Conn, Andrew R., Nicholas IM Gould, and Philippe L. Toint. "Trust region methods." Society for Industrial and Applied Mathematics, 2000.

\bibitem{berghen2004levenberg}
Berghen, Frank Vanden. "Levenberg-Marquardt algorithms vs trust region algorithms." IRIDIA, Université Libre de Bruxelles 1 (2004).

\bibitem{ngia2000efficient}
Ngia, Lester SH, and Jonas Sjoberg. "Efficient training of neural nets for nonlinear adaptive filtering using a recursive Levenberg-Marquardt algorithm." IEEE Transactions on Signal Processing 48.7 (2000): 1915-1927.

\bibitem{asirvadam2002parallel}
V. S. Asirvadam, S. F. McLoone and G. W. Irwin, "Parallel and separable recursive Levenberg-Marquardt training algorithm," Proceedings of the 12th IEEE Workshop on Neural Networks for Signal Processing, 2002.

\bibitem{auger2012making}
Auger, François, E. Chassande-Mottin, and Patrick Flandrin. "Making reassignment adjustable: the Levenberg-Marquardt approach." 2012 IEEE International Conference on Acoustics, Speech and Signal Processing (ICASSP). IEEE, 2012.

\bibitem{nguyen2013combining}
Nguyen, Lien B., et al. "Combining genetic algorithm and Levenberg-Marquardt algorithm in training neural network for hypoglycemia detection using EEG signals." 2013 35th annual international conference of the IEEE Engineering in Medicine and Biology Society (EMBS). IEEE, 2013.

\bibitem{yu2018levenberg}
Yu, Hao, and Bogdan M. Wilamowski. "Levenberg–Marquardt training." Intelligent Systems. CRC Press, 2018. 12-1.

\bibitem{gavin2019levenberg}
Gavin, Henri P. "The Levenberg-Marquardt algorithm for nonlinear least squares curve-fitting problems." Department of Civil and Environmental Engineering, Duke University (2019): 1-19.

\bibitem{sarkka2020levenberg}
Särkkä, Simo, and Lennart Svensson. "Levenberg-Marquardt and line-search extended Kalman smoothers." ICASSP 2020-2020 IEEE International Conference on Acoustics, Speech and Signal Processing (ICASSP). IEEE, 2020.

\bibitem{cheung2006constrained}
Cheung, Ka Wai, et al. "A constrained least squares approach to mobile positioning: algorithms and optimality." EURASIP Journal on Advances in Signal Processing 2006 (2006): 1-23.

\bibitem{lopes_2008}
Lopes, Cassio G., and Ali H. Sayed. "Diffusion least-mean squares over adaptive networks: Formulation and performance analysis." IEEE Transactions on Signal Processing 56.7 (2008): 3122-3136.

\bibitem{keaktos_2011}
Kekatos, Vassilis, and Georgios B. Giannakis. "Sparse Volterra and polynomial regression models: Recoverability and estimation." IEEE Transactions on Signal Processing 59.12 (2011): 5907-5920.

\bibitem{zhu_2011}
Zhu, Hao, Geert Leus, and Georgios B. Giannakis. "Sparsity-cognizant total least-squares for perturbed compressive sampling." IEEE Transactions on Signal Processing 59.5 (2011): 2002-2016.

\bibitem{sahu_2016}
Sahu, Anit Kumar, et al. "Distributed constrained recursive nonlinear least-squares estimation: Algorithms and asymptotics." IEEE Transactions on Signal and Information Processing over Networks 2.4 (2016): 426-441.

\bibitem{fu_2015}
Fu, Xingang, et al. "Training recurrent neural networks with the Levenberg–Marquardt algorithm for optimal control of a grid-connected converter." IEEE Transactions on Neural Networks and Learning Systems 26.9 (2014): 1900-1912.

\bibitem{xu_2016}
Xu, Fan, Fenfang Li, and Yuanqing Wang. "Modified Levenberg–Marquardt-based optimization method for LiDAR waveform decomposition." Ieee Geoscience and Remote Sensing Letters 13.4 (2016): 530-534.

\bibitem{meng_2019}
Meng, Hao, et al. "Indoor positioning of RBF neural network based on improved fast clustering algorithm combined with LM algorithm." IEEE Access 7 (2018): 5932-5945.

\bibitem{ok_2018}
Ok, Ali Ozgun, and Asli Ozdarici-Ok. "Combining orientation symmetry and LM cues for the detection of citrus trees in orchards from a digital surface model." IEEE Geoscience and Remote Sensing Letters 15.12 (2018): 1817-1821.

\bibitem{dempster1977maximum}
Dempster, Arthur P., Nan M. Laird, and Donald B. Rubin. "Maximum likelihood from incomplete data via the EM algorithm." Journal of the Royal Statistical Society: Series B (Methodological) 39.1 (1977): 1-22.

\bibitem{jacobson2007expanded}
Jacobson, Matthew W., and Jeffrey A. Fessler. "An expanded theoretical treatment of iteration-dependent Majorize-Minimize algorithms." IEEE Transactions on Image Processing 16.10 (2007): 2411-2422.

\bibitem{lange1995gradient}
Lange, Kenneth. "A gradient algorithm locally equivalent to the EM algorithm." Journal of the Royal Statistical Society: Series B (Methodological) 57.2 (1995): 425-437.

\bibitem{schifano2010majorization}
Schifano, Elizabeth D., Robert L. Strawderman, and Martin T. Wells. "Majorization-Minimization algorithms for nonsmoothly penalized objective functions." Electronic Journal of Statistics 4 (2010): 1258-1299.

\bibitem{chouzenoux2017stochastic}
Chouzenoux, Emilie, and Jean-Christophe Pesquet. "A stochastic Majorize-Minimize subspace algorithm for online penalized least squares estimation." IEEE Transactions on Signal Processing 65.18 (2017): 4770-4783.

\bibitem{marnissi2020majorize}
Marnissi, Yosra, et al. "Majorize–Minimize Adapted Metropolis–Hastings Algorithm." IEEE Transactions on Signal Processing 68 (2020): 2356-2369.

\bibitem{lange2021nonconvex}
Lange, Kenneth, et al. "Nonconvex Optimization via MM Algorithms: Convergence Theory." arXiv preprint arXiv:2106.02805 (2021).

\bibitem{fest2021stochastic}
Fest, Jean-Baptiste, and Emilie Chouzenoux. "Stochastic Majorize-Minimize Subspace Algorithm with Application to Binary Classification." 29th European Signal Processing Conference (EUSIPCO 2021). 2021.

\bibitem{sun2016majorization}
Sun, Ying, Prabhu Babu, and Daniel P. Palomar. "Majorization-Minimization algorithms in signal processing, communications, and machine learning." IEEE Transactions on Signal Processing 65.3 (2016): 794-816.

\bibitem{diamond2016cvxpy}
Diamond, Steven, and Stephen Boyd. "CVXPY: A Python-embedded modeling language for convex optimization." The Journal of Machine Learning Research 17.1 (2016): 2909-2913.

\bibitem{livingstone_steven_r_2018_1188976}
Livingstone, Steven R., and Frank A. Russo. "The Ryerson Audio-Visual Database of Emotional Speech and Song (RAVDESS): A dynamic, multimodal set of facial and vocal expressions in North American English." PloS one 13.5 (2018): e0196391.

\end{thebibliography}
\end{document}